\documentclass[lettersize,journal]{IEEEtran}
\usepackage{amsmath,amsfonts}
\usepackage{array}
\usepackage[caption=false,font=normalsize,labelfont=sf,textfont=sf]{subfig}
\usepackage{textcomp}
\usepackage{stfloats}
\usepackage{url}
\usepackage{mathtools}
\usepackage{verbatim}
\usepackage{multirow}
\usepackage{graphicx}
\usepackage{booktabs}
\usepackage{colortbl}
\usepackage{xcolor}
\usepackage{hyperref}
\definecolor{V}{RGB}{21,137,139}
\definecolor{X}{RGB}{234,120,60}
\usepackage{cite}
\usepackage{algorithm}
\usepackage{algpseudocode}
\usepackage[algo2e]{algorithm2e}
\begin{document}

\title{Fine-Grained VLM Fine-tuning via Latent Hierarchical Adapter Learning}

\author{Yumiao Zhao, Bo Jiang*, Yuhe Ding, Xiao Wang, Jin Tang, Bin Luo 
\thanks{

The authors are with the Information Materials and Intelligent
Sensing Laboratory of Anhui Province, Anhui Provincial Key Laboratory of
Multimodal Cognitive Computation, School of Computer Science and Technology, Anhui University

*Corresponding author

}
\thanks{Manuscript received April 19, 2021; revised August 16, 2021.}}

\markboth{Journal of \LaTeX\ Class Files,~Vol.~14, No.~8, August~2021}%
{Shell \MakeLowercase{\textit{et al.}}: A Sample Article Using IEEEtran.cls for IEEE Journals}

\IEEEpubid{0000--0000/00\$00.00~\copyright~2021 IEEE}
\maketitle
\begin{abstract}
Adapter-based approaches have garnered attention for fine-tuning pre-trained Vision-Language Models (VLMs) on few-shot classification tasks.  
 These methods strive to develop a lightweight module that better aligns visual and (category) textual representations, thereby enhancing performance on downstream few-shot learning tasks.  
However, existing adapters generally learn/align (category) textual-visual modalities via explicit spatial proximity in the underlying embedding space, which i)  fails to capture the inherent one-to-many associations between categories and image samples and ii) struggles to establish accurate associations between the unknown categories and images. 
To address these issues, inspired by recent works on hyperbolic learning, we 
develop a novel Latent Hierarchical Adapter (LatHAdapter) for fine-tuning VLMs on downstream few-shot classification tasks.  
The core of LatHAdapter is to exploit the \emph{latent semantic hierarchy} of downstream training data and employ it to provide \emph{richer, fine-grained guidance} for the adapter learning process.  
Specifically, LatHAdapter first introduces some learnable `attribute' prompts as the bridge to align categories and images. 
Then, it projects the categories, attribute prompts, and images within each batch in a hyperbolic space, and employs hierarchical regularization to learn the latent semantic hierarchy of them, thereby fully modeling the inherent one-to-many associations among categories, learnable attributes, and image samples. Extensive experiments on four challenging few-shot tasks show that the proposed LatHAdapter consistently outperforms many other fine-tuning approaches, particularly in adapting known classes and generalizing to unknown classes. Code is available at: \href{Code}{https://github.com/zhaoym55/HyperbolicAdapter.git}

\end{abstract}


\section{Introduction}
\IEEEPARstart{R}{ecently}, Vision-Language Models (VLMs)~\cite{radford2021learning,jia2021scaling} have demonstrated remarkable cross-domain and zero-shot generalization ability. Trained on large-scale image-text paired data, these models can align visual and textual modalities in a shared embedding space. 
To efficiently transfer pre-trained VLMs to downstream tasks, Parameter-Efficient Fine-Tuning (PEFT) methods are widely studied, focusing on fine-tuning only a small number of parameters with few-shot labeled training data. 

\begin{figure}[!h]
  \centering
  \includegraphics[width=0.9\linewidth]{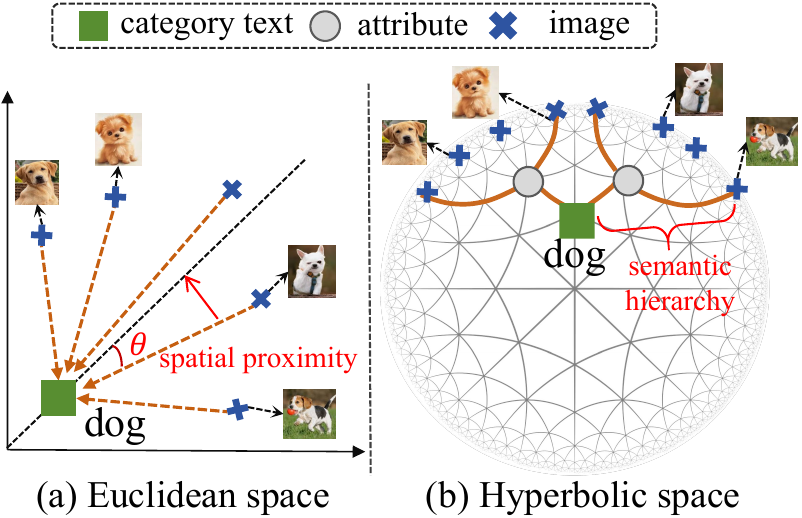}
  \caption{ Previous methods typically use Euclidean distance to enforce explicit spatial proximity constraints in the embedding space. The matching text–image sample pairs are close, while the non-matching pairs are apart(left). In contrast, LatHAdapter (right) employs hyperbolic geometry to construct a category$\rightarrow$attribute$\rightarrow$image semantic hierarchy, enabling one-to-many alignment from text categories to image samples, thereby enhancing cross-modal representation learning.} 
\label{motivation}
\end{figure}

Overall, there are three types of  PEFT approaches for VLMs, including prompt learning~\cite{hu2024comma,khattak2023self,jia2022visual,zhou2022learning}, low-rank factorization~\cite{hayou2024lora,hu2022lora,valipour2022dylora,shi2024reslora}, and adapter tuning~\cite{zhang2022tip,yu2023task,gao2024clip,li2024graphadapter}. Prompt learning methods aim to introduce learnable prompt vectors to fine-tune pre-trained VLMs, enabling effective adaptation to specific tasks. 
For instance, CoOp~\cite{zhou2022learning} integrates continuous learnable vectors before class names to refine the textual features. In contrast, VPT~\cite{jia2022visual} focuses on improving the image encoder by introducing learnable visual prompts to capture task-specific knowledge. To jointly refine the text and image encoders, MaPLe~\cite{khattak2023maple} introduces branch-aware hierarchical prompts into two encoders, facilitating effective adaptation to downstream tasks. Low-rank factorization works introduce low-rank matrices to update the model weights, without modifying the original model architecture. For example, Low-Rank Adaptation (LoRA)~\cite{hu2022lora} introduces trainable low-rank matrices into the original model to enable parameter-efficient fine-tuning. LoRA+~\cite{hayou2024lora} builds on LoRA by introducing proportional learning rates for low-rank matrices A and B, facilitating more efficient learning of task-specific features. \IEEEpubidadjcol To further enhance task-specific performance, LoRA-Dash~\cite{si2024unleashing} first identifies task-specific directions and maximizes their impact during the fine-tuning process. Unlike previous approaches, adapter tuning methods focus on introducing lightweight modules to adapt VLMs for multiple tasks. Such as Tip-Adapter~\cite{zhang2022tip} constructs a query-key cache module from training samples and utilizes prior knowledge to optimize the image encoder. Graph-adapter~\cite{li2024graphadapter} designs textual and visual knowledge graphs to model inter-modality relationships, refining the textual embeddings. MMA~\cite{seputis2024multi} proposes a multi-modal adapter with both modality-specific and shared projection layers, enabling cross-modal interaction and enhancing text–image alignment during fine-tuning.

It is known that one important aspect of VLM adapter approaches is to develop a lightweight module to further improve the alignment between visual and (category)  textual representations on the downstream training data, thereby enhancing the performance of pre-trained VLMs on downstream few-shot classification tasks. 
From this perspective, after reviewing previous works, we observe that existing adapter-based few-shot learning approaches generally exhibit two main limitations.  
%
\textbf{First}, 
existing works generally conduct (category) textual-visual modality alignment learning via explicit spatial proximity constraints in the underlying embedding space, i.e., the representations of category and matching images are pulled together
~\cite{gao2024clip,hao2025task,li2025atprompt}, as shown in Fig.~\ref{motivation} (left). 
Obviously, such spatial proximity constraints are not reasonable for few-shot classification tasks. 
%
It is more reliable to model the relationships between category and image samples via the hierarchical one-to-many associations~\cite{desai2023hyperbolic,ramasinghe2024accept}, as illustrated in Fig.~\ref{motivation} (right). 
%
%
\textbf{Second}, 
previous approaches generally describe the class-specific textual features by either employing hand-crafted text prompts~\cite{seputis2024multi} (e.g., `a photo of a [class]'), or generating diverse class descriptions via large language models (LLMs)~\cite{tang2024amu}. 
However, these explicit textual descriptors of categories usually restrict existing fine-tuning methods to align images with predefined categories during training, limiting their ability to generalize to unknown categories. 

To address these issues, motivated by the advantages of deep hyperbolic learning~\cite{kim2023hier,Compositional2024,desai2023hyperbolic},  
in this paper, we 
develop a novel Latent Hierarchical Adapter (LatHAdapter) for fine-tuning VLMs on few-shot classification tasks. 
The core aspect of our LatHAdapter is to first learn the \textbf{latent semantic hierarchy} of downstream training data and then fully exploit it to provide \textbf{richer, fine-grained guidance} for the fine-tuning problem. To achieve this goal, LatHAdapter first introduces some learnable `attribute' prompts and designs an Attribute-aware Text Refiner (ATR) module to bridge the gap between categories and images.  
ATR is implemented with no annotation but by learning some universal  `attributes’ adaptively.
%
Then, LatHAdapter embeds the categories, learnable attributes, and image samples concurrently in a unified hyperbolic space and employs a hierarchical regularization 
to learn the \emph{compact representations} and \emph{latent semantic hierarchy} among categories, attributes, and images for the downstream training data. This enables LatHAdapter to fully capture the one-to-many associations between categories and attributes, as well as between attributes and images to provide richer, fine-grained guidance for fine-tuning tasks. 
Finally, we jointly optimize the downstream label prediction
and regularization loss functions to optimize LatHAdapter. 

We note that hyperbolic learning has been exploited for pre-training or fine-tuning pre-trained models in previous work MERU~\cite{desai2023hyperbolic}, HyperCLIP~\cite{Open2025}, and HySAC~\cite{poppi2025hyperbolic}. 
MERU leverages the entailment cone to model the image-text hierarchical structure in the pre-training CLIP model. 
HySAC~\cite{poppi2025hyperbolic} also
employs entailment
loss functions to model hierarchical and asymmetrical
relations between safe and unsafe image-text pairs for unsafe content retrieval. 
HyperCLIP~\cite{Open2025} proposes to adjust the hyperbolic radius of textual embeddings to facilitate better alignment between text and pixels for semantic segmentation tasks.  
In contrast, this paper mainly focuses on adapting VLMs on the downstream few-shot classification tasks. 
%
Unlike previous works that generally leverage a two-level hierarchy (text $\rightarrow$ image/pixels) in hyperbolic space, the proposed LatHAdapter further introduces learnable attribute prompts to learn the latent hierarchical {triplet} representation (category $\rightarrow$ attribute $\rightarrow$ image) inherent in each batch of downstream data. 
\textbf{This encourages fine-grained VLM fine-tuning and also enhances generalization to unknown categories.} 
Moreover, LatHAdapter employs two triplet regularization losses to further encourage \emph{compact} representations for categories, attributes, and images. 

In summary, the main contributions of the proposed method are as follows:

\begin{itemize}
\item 
We present a novel Latent Hierarchical Adapter (LatHAdapter) for fine-tuning
VLMs on few-shot classification tasks. It  
can fully capture the semantic hierarchy (category $\rightarrow$ attribute $\rightarrow$ image) of downstream training data in each batch for adapter learning.

\item We design an Attribute-aware Text Refiner (ATR) module to integrate the attribute information into textual embeddings, enhancing fine-grained textual representations and improving generalization to new classes. It provides rich and fine-grained guidance for adapter learning.

\item 
LatHAdapter provides a plug-and-play scheme. By integrating LatHAdapter into several pre-trained VLMs, we can achieve consistently improved performance over the state-of-the-art methods on few-shot classification tasks. 

\end{itemize}

Extensive experiments on four challenging tasks, including base-to-new class generalization, cross-dataset generalization, domain generalization, and few-shot image classification, demonstrate the effectiveness of LatHAdapter. 
The rest of the paper is organized as follows. In Section~\ref{relatedwork}, we review prior works on PEFT for VLMs and  Hyperbolic learning. Section~\ref{Preliminaries} introduces the preliminaries, including an overview of vision-language models (VLMs) and the theoretical foundation of the Poincaré ball model. In Section~\ref{method}, we present our LatHAdapter in detail. In Section~\ref{experiment}, we integrate the LatHAdapter into three representative PEFT methods and evaluate their performance on various tasks.

\section{Related Work}
\label{relatedwork}

\subsection{Parameter-Efficient Fine-Tuning (PEFT)}
Vision Language Models (VLMs)~\cite{Chen_2021,radford2021learning, Vilbert2019}, leveraging large-scale text-image pairs for training, demonstrate remarkable zero-shot generalization capabilities in image recognition tasks. To adapt pre-trained VLMs for downstream tasks, traditional fine-tuning methods update all parameters. However, these methods are prone to overfitting in low-data scenarios and require high computational costs. To address this issue, Parameter-Efficient Fine-Tuning (PEFT) techniques are proposed, including prompt learning, low-rank factorization, and adapter tuning. Unlike hand-crafted text prompts used in CLIP, prompt learning methods~\cite{khattak2023maple,khattak2023self,Conceptual2024,hu2024comma} focus on optimizing continuous prompt vectors to adapt pre-trained models for downstream tasks. CoCoOp~\cite{zhou2022conditional} introduces learnable text prompts and utilizes visual sample features to enhance the representation of textual features. TextRefiner~\cite{xie2024textrefiner} refines text prompts by incorporating fine-grained visual concepts extracted from local visual tokens. DPC~\cite{li2025dpc} proposes a dual-prompt framework with a weighting-decoupling strategy to balance task-specific discrimination and generalization to unknown categories. ATPrompt~\cite{li2025atprompt} introduces universal attribute descriptions into the text templates, thereby improving alignment between images and unknown categories.
Low-rank factorization methods~\cite{hu2022lora,valipour2022dylora,zhao2024hegraphadapter} insert trainable low-rank matrices into the model. During inference, the low-rank matrices can be merged back into the original model, preserving both the model architecture and inference speed. Dora~\cite{mao2024dora} decomposes LoRA layers into rank-one components and dynamically prunes them during training based on their task-specific importance. AdaLoRA~\cite{zhang2023adalora}  adaptively allocates the parameter budget for weight matrices based on their importance, thereby enhancing fine-tuning efficiency.
Different from these methods, adapter tuning methods~\cite{gao2024clip, zhang2022tip, seputis2024multi} introduce lightweight modules in pre-trained VLMs to adapt them for specific tasks. Existing adapter-based approaches typically refine textual features~\cite{yu2023task,li2024graphadapter}, visual features~\cite{zhang2022tip,Proto2024}, or joint optimizing both textual and visual features~\cite{zhao2024hegraphadapter,guo2025mmrl,seputis2024multi}. For example, CLIP-Adapter~\cite{gao2024clip} inserts a lightweight bottleneck layer in the image and text encoders, enabling the network to learn task-specific knowledge. Meta-Adapter~\cite{song2023meta} introduces a residual-style adapter that refines textual embeddings by injecting the few-shot visual knowledge into textual features. Mmrl~\cite{guo2025mmrl} introduces a shared, learnable representation space and projects its vectors into text and image representation tokens, thereby enabling more effective multi-modal interactions. Hegraphadapter~\cite{zhao2024hegraphadapter} introduces a Heterogeneous Graph Adapter that models the relationships between textual and visual modalities, facilitating the joint optimization of both the text and image encoders.
These methods align textual and visual modalities via explicit spatial proximity constraints in Euclidean space. However, they overlook the inherent semantic hierarchy between modalities and fail to model one-to-many correspondences between text categories and image samples.

\subsection{Hyperbolic Learning}

Hyperbolic learning has been successfully applied in various domains, including few-shot learning~\cite{2025Hyperbolic,2021Curvature} and vision-language models~\cite{Compositional2024,Open2025}. Unlike Euclidean space, hyperbolic space naturally supports the representation of hierarchical and tree-like structures, due to its exponential volume growth with respect to radius~\cite{krioukov2010hyperbolic,cannon1997hyperbolic}. Among various isometric models of hyperbolic space, the Poincaré ball model~\cite{nickel2017poincare,ganea2018hyperbolic} and the Lorentz model~\cite{nickel2018learning,ramasinghe2024accept} are two commonly adopted models. For example, Pal et al.~\cite{Compositional2024} leverage the Lorentz model to construct hierarchical compositional relations of images, image boxes, and their textual descriptions, thereby enhancing alignment between textual and visual modalities. Peng et al.~\cite{Open2025} fine-tune CLIP in hyperbolic space by scaling the hyperbolic radius of text embeddings, enabling hierarchical alignment for open-vocabulary segmentation tasks. Yang et al.~\cite{2025Hyperbolic} capture hierarchical relations among images from different domains, realizing more effective generalization in cross-domain few-shot learning. Poppi et al.~\cite{poppi2025hyperbolic} employ hyperbolic entailment loss to model the hierarchical and asymmetrical relations between safe and unsafe image-text pairs. Hu et al.~\cite{hu2024rethinking} exploit hyperbolic constraints between prototypes and instances to improve domain alignment and feature discrimination. In this work, we construct a category$\rightarrow$attribute$\rightarrow$image hierarchical structure among text classes, image instances, and attribute prompts via the Poincaré ball model, thereby enabling richer and fine-grained guidance during fine-tuning.

\section{Preliminaries}
\label{Preliminaries}

\subsection{Poincaré Ball Space}
Hyperbolic space is defined as a non-Euclidean Riemannian manifold with negative curvature. Due to its exponential volume growth, hyperbolic space can facilitate more efficient embedding of hierarchical and tree-like data structures. This property makes it well-suited for modeling complex semantic hierarchies in vision-language tasks. Among various representations of hyperbolic geometry, the Poincaré ball model~\cite{sarkar2011low,nickel2017poincare} is widely used for various tasks that involve gradient-based optimization. The Poincaré ball model is defined as ($\mathbb{D}_c^n,g^\mathbb{D}$), where $\mathbb{D}_c^n=\{ \boldsymbol{x}\in\mathbb{R}^n:c||\boldsymbol{x}||<1\}$ and Riemannian metric $g^\mathbb{D}=\lambda_c^2g^E$. The hyperparameter $c$ controls the curvature of the ball. The $\lambda_c=\frac{2}{1-c||\boldsymbol{x}||^2}$ represents the conformal factor.
To project vector $\boldsymbol{x}$ from Euclidean space $\mathbb{R}^n$ to hyperbolic space $\mathbb{D}^n$, we adopt the Exponential Map function~\cite{kim2023hier}, which is defined as:
\begin{equation}
\mathrm{exp}_{0}^c(\boldsymbol{x})=\mathrm{tanh}(\sqrt{c}||\boldsymbol{x}||)\frac{\boldsymbol{x}}{\sqrt{c}||\boldsymbol{x}||}.
\label{expmap}
\end{equation}
Due to the non-Euclidean properties of hyperbolic space, standard vector operations must be redefined using gyrovector spaces~\cite{Abraham2008}. For the vectors $\boldsymbol{u}\in \mathbb{D}_c^n$ and $\boldsymbol{z}\in \mathbb{D}_c^n$, we use Möbius addition to realize the addition operation, which is formulated as:

\begin{equation}
\boldsymbol{u} \oplus_c \boldsymbol{z} = 
\frac{
(1 + 2c \langle \boldsymbol{u}, \boldsymbol{z} \rangle + c \|\boldsymbol{z}\|^2)\boldsymbol{u} + (1 - c \|\boldsymbol{u}\|^2)\boldsymbol{z}
}{
1 + 2c \langle \boldsymbol{u}, \boldsymbol{z} \rangle + c^2 \|\boldsymbol{u}\|^2 \|\boldsymbol{z}\|^2
}.
\end{equation}
The hyperbolic distance between $\boldsymbol{u}$ and $\boldsymbol{z}$ in Poincaré ball is defined as follows:
\begin{equation}
d_H(\boldsymbol{u}, \boldsymbol{z}) = \frac{2}{\sqrt{c}} \, \mathrm{arctanh}\left( \sqrt{c} \left\| -\boldsymbol{u} \oplus_c \boldsymbol{z} \right\| \right).
\label{distance}
\end{equation}
When the curvature $c$ is set to zero, the hyperbolic distance becomes equivalent to the Euclidean distance.

\subsection{Vision-Language Models}
Large-scale foundation Vision-language models (VLMs), such as CLIP, show remarkable zero-shot generalization capabilities. CLIP is pretrained on a dataset of a billion image-text pairs and employs a contrastive learning loss to align the visual and textual modalities in a shared embedding space. The CLIP model includes two encoders: a text encoder $\mathcal{T}(\cdot)$ and an image encoder $\mathcal{V}(\cdot)$. In this paper, we adopt a vision transformer (ViT) as the backbone for the image encoder. Given an image matrix $\mathbf{X}^v$, we obtain a visual embedding as $\boldsymbol{v}=\mathcal{V}(\mathbf{X}^v)$. For each class, we construct a text description $\mathbf{X}^t$ by inserting the class name into a predefined text template (e.g., ``A photo of a [class]"), and obtain the textual embedding as $\boldsymbol{t}=\mathcal{T}(\mathbf{X}^t)$. Finally, the classification probability for image $\mathbf{X}^v$ is calculated as follows:
\begin{equation}
P(y=i|\mathbf{X}^v)=\frac{\mathrm{exp}(\mathrm{cos}(\boldsymbol{v},\boldsymbol{t}_i)/\tau)}{\sum_{j=1}^{C}\mathrm{exp}(\mathrm{cos}(\boldsymbol{v},\boldsymbol{t}_j)/\tau)},
\end{equation}
where $\mathrm{cos}(\cdot)$ is the cosine similarity function, and $\tau$ denotes the temperature parameter. The $\boldsymbol{t}_i$ indicates the text embedding corresponding to the $i$-th class, and $C$ denotes the total number of classes in the current task.

\section{Latent Hierarchical Adapter}
\label{method}

\begin{figure*}[!h]
  \centering
  \includegraphics[width=1\linewidth]{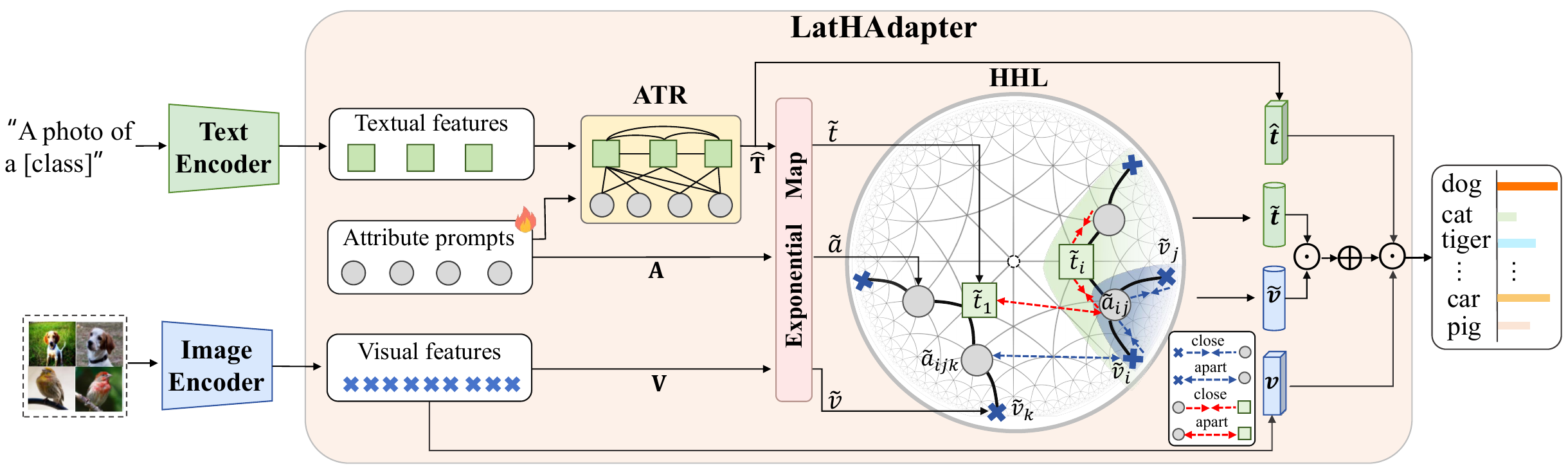}
  \caption{ The framework of the proposed LatHAdapter. We first introduce a set of learnable attribute prompts, which aim to construct accurate associations between known and unknown classes. We then design an Attribute-aware Text Refiner (ATR) module to enhance fine-grained textual representations and improve generalization to unknown classes. Subsequently, we use the Exponential Map function to project the text categories, learnable attributes, and image samples in a unified hyperbolic space. To further exploit latent semantic hierarchies of downstream training data, we introduce Hyperbolic Hierarchical Learning (HHL), which models a category$\rightarrow$attribute$\rightarrow$image semantic hierarchy and establishes one-to-many associations between text categories and image samples. Finally, we jointly optimize a contrastive loss and hierarchical regularization loss for adapter learning. 
  } 
\label{network}
\end{figure*}

Although pre-trained VLMs achieve remarkable performance, existing works~\cite{liang2022mind} reveal that the gap between visual and textual modalities still exists, and it is necessary to deal with the gap between different modalities on the fine-tuning stage. 
As discussed in the Introduction section, existing Adapter methods generally fail to model the hierarchical one-to-many correspondences between textual categories and visual modalities. 
To address this issue, in this section, we propose a novel Latent Hierarchical Adapter (LatHAdapter) that fully exploits the latent semantic hierarchies of downstream training data to provide richer fine-grained guidance for adapter learning. 
LatHAdapter comprises two main modules: i) Attribute-aware Text Refiner (ATR) module that refines textual representation for categories through learnable `attribute' prompts, and ii) Hyperbolic Hierarchical Learning (HHL) that captures latent semantic hierarchy and fully models the inherent one-to-many associations among categories, attributes, and image samples. The overall framework is illustrated in Fig.~\ref{network}. 

\subsection{Attribute-aware Text Refiner}
Given a batch of image samples, we extract visual embedding $\mathbf{V}=\{\boldsymbol{v}_1,\boldsymbol{v}_2,\cdots,\boldsymbol{v}_B\}$ using the image encoder of pre-trained VLM model. For the downstream task, we obtain textual embeddings as $\mathbf{T}=\{\boldsymbol{t}_1,\boldsymbol{t}_2,\cdots,\boldsymbol{t}_{C}\}$ via the text encoder. 
We then introduce learnable `attribute' prompts $\mathbf{A}=\{\boldsymbol{a}_1,\boldsymbol{a}_2,\cdots,\boldsymbol{a}_{N}\}\in \mathbb{R}^{N\times d}$ as intermediate semantic representations to bridge the gap between visual and textual modalities, where $N$ is the number of prompt vectors and $d$ denotes the dimension.  
To effectively achieve the interaction between categories $\mathbf{T}$ and attributes $\mathbf{A}$, we design an Attribute-aware Text Refiner (ATR) module that employs a self-attention (SA)  mechanism to refine category representation.
Formally, let 
$\mathbf{F}=\{\mathbf{T},\mathbf{A}\}=\{\boldsymbol{f}_1,\boldsymbol{f}_2,\cdots,\boldsymbol{f}_{C+N}\}$, 
we first compute the affinities $\mathbf{S}_{ij}$ between $\boldsymbol{f}_i$ and $\boldsymbol{f}_j$ to capture the correlation between the downstream task classes and attribute prompts. 
Then, we utilize a message aggregation function to refine each category representation. 
In summary, ATR learn the refined textual features $\hat{\mathbf{T}}$ as follows:
\begin{align}
\label{down}
&\mathbf{S}_{ij}= \mathrm{cos}(\mathbf{f}_i,\mathbf{f}_j),\\
&\hat{\mathbf{F}}= \mathbf{F}+\beta(\mathbf{D}^{-\frac{1}{2}}\mathbf{S}\mathbf{D}^{-\frac{1}{2}}\mathbf{F}),\\
&\hat{\mathbf{T}}=\hat{\mathbf{F}}[0:C,:],
\end{align}
where $\beta$ is the hyperparameter. $\mathbf{D}=\mathrm{diag}(\sum_{j}^{C}\mathbf{S}_{ij})$ is the degree matrix. $C$ denotes the number of classes. 

\subsection{Hyperbolic Hierarchical Learning }
To explore latent semantic hierarchies of downstream training data, we project categories, attributes, and images into a hyperbolic space and design a hierarchical learning regularization to learn their representations.  
Specifically, we leverage the Poincaré disk model, whose exponential geometry can naturally support the semantic hierarchical modeling~\cite{nickel2017poincare}. 
In this hyperbolic space, category concepts are positioned near the origin, intermediate-level attributes are located at mid-regions, and fine-grained visual features are placed closer to the boundary. Leveraging this hierarchical geometry, the Hyperbolic Hierarchical Learning (HHL) module effectively learns transferable attribute representations guided by visual and textual information from downstream tasks. Also, it models the one-to-many associations between categories and attributes, as well as between attributes
and images to provide fine-grained guidance for fine-tuning tasks.



\subsubsection{\textbf{Hyperbolic Embedding}}
We project the refined category embedding $\hat{\boldsymbol{t}}$, visual embedding $\boldsymbol{v}$, and attribute embedding $\boldsymbol{a}$ into the hyperbolic space as, 
\begin{align}
\label{down}
\tilde{\boldsymbol{t}}=\mathrm{exp}_{0}^c(\hat{\boldsymbol{t}}),\,\, \tilde{\boldsymbol{v}}= \mathrm{exp}_{0}^c(\boldsymbol{v}),\,\, \tilde{\boldsymbol{a}}=\mathrm{exp}_{0}^c(\boldsymbol{a}),
\end{align}
where $\text{exp}_{0}^c(\cdot)$ denotes the Exponential Map function, which is defined in Eq. (\ref{expmap}) and $c$ is the curvature of the ball. 

\subsubsection{\textbf{Image-Attribute Hierarchical Learning}}
We first model image-attribute hierarchical relationships between image samples and attribute prompts. For image samples, we construct a series of triplets $\{\tilde{\boldsymbol{v}}_i, \tilde{\boldsymbol{v}}_j, \tilde{\boldsymbol{v}}_k\}$ as follows:
\begin{align}
\{\tilde{\boldsymbol{v}}_i, \tilde{\boldsymbol{v}}_j, \tilde{\boldsymbol{v}}_k\}=(\tilde{\boldsymbol{v}}_j\in \mathbf{N}_K(\tilde{\boldsymbol{v}}_i))\wedge (\tilde{\boldsymbol{v}}_k\notin \mathbf{N}_K(\tilde{\boldsymbol{v}}_i)\},
\end{align}
where $(\tilde{\boldsymbol{v}}_i, \tilde{\boldsymbol{v}}_j)$ denotes a positive pair, and $(\tilde{\boldsymbol{v}}_i, \tilde{\boldsymbol{v}}_k)$ is a negative pair. $\mathbf{N}_K(\cdot)$ represents the k-nearest neighbors. 
To identify the shared hierarchical attribute $\boldsymbol{a}_{i,j}$ as the lowest common ancestor (LCA)~\cite{kim2023hier} of a positive image pair $(\tilde{\boldsymbol{v}}_i, \tilde{\boldsymbol{v}}_j)$, we use a probability function $\pi_{ij}(\cdot)$ to estimate the likelihood of each attribute prompt $\tilde{\boldsymbol{a}}\in\tilde{\mathbf{A}}$ being the LCA, which is calculated as:
\begin{equation}
\begin{aligned}
& \pi_{ij}(\tilde{\boldsymbol{a}})=\mathrm{exp}(-\mathrm{max}(d_H(\tilde{\boldsymbol{v}}_i, \tilde{\boldsymbol{a}}),d_H(\tilde{\boldsymbol{v}}_j, \tilde{\boldsymbol{a}}))), \\
& \tilde{\boldsymbol{a}}_{ij}=\mathrm{\arg\max}(\pi_{ij}(\tilde{\boldsymbol{a}})+g),
 \label{psotive}
 \end{aligned}
 \end{equation}
where $d_H(\cdot)$ denotes the hyperbolic distance function, as defined in Eq. (\ref{distance}). $g$ is Gumbel noise sampled from $\mathrm{Gumbel}(0,1)$ used to prevent local optima. Similarly, the common hierarchical attribute  $\tilde{\boldsymbol{a}}_{ijk}$ can be obtained as the LCA of the entire triplet $\{\tilde{\boldsymbol{v}}_i, \tilde{\boldsymbol{v}}_j, \tilde{\boldsymbol{v}}_k\}$ as:
\begin{equation}
\begin{aligned}
& \pi_{ijk}(\tilde{\boldsymbol{a}})=\mathrm{exp}(-\mathrm{max}(d_H(\tilde{\boldsymbol{v}}_i, \tilde{\boldsymbol{a}}),d_H(\tilde{\boldsymbol{v}}_j, \tilde{\boldsymbol{a}}),d_H(\tilde{\boldsymbol{v}}_k, \tilde{\boldsymbol{a}}))),\\
& \tilde{\boldsymbol{a}}_{ijk}=\mathrm{\arg\max}(\pi_{ijk}(\tilde{\boldsymbol{a}})+g).
 \label{entire}
 \end{aligned}
 \end{equation}
To ensure that the positive image pairs are closing, while the negative images are away from them, we introduce a hierarchical regularization loss~\cite{kim2023hier} for the image-attribute structure, which is formulated as:
\begin{equation}
\begin{split}
\mathcal{L}_{v-a}& =\mathrm{ReLU}[d_H(\tilde{\boldsymbol{v}}_i, \tilde{\boldsymbol{a}}_{ij})-d_H(\tilde{\boldsymbol{v}}_i,  \tilde{\boldsymbol{a}}_{ijk})+\sigma]\\
& +\mathrm{ReLU}[d_H(\tilde{\boldsymbol{v}}_j, \tilde{\boldsymbol{a}}_{ij})-d_H(\tilde{\boldsymbol{v}}_j, \tilde{\boldsymbol{a}}_{ijk})+\sigma]\\
& +\mathrm{ReLU}[d_H(\tilde{\boldsymbol{v}}_k, \tilde{\boldsymbol{a}}_{ijk})-d_H(\tilde{\boldsymbol{v}}_k, \tilde{\boldsymbol{a}}_{ij})+\sigma],
\label{av_reg}
\end{split}
\end{equation}
where $\sigma$ is a margin hyperparameter. The $\mathcal{L}_{v-a}$ facilitates the positive pair $(\tilde{\boldsymbol{v}}_i, \tilde{\boldsymbol{v}}_j)$ should be closer to their common hierarchical attribute $\tilde{\boldsymbol{a}}_{i,j}$. The negative image sample $\tilde{\boldsymbol{v}}_k$ is encouraged to move away from $\tilde{\boldsymbol{a}}_{i,j}$ to a different branch of the hierarchy. As a result,  $\tilde{\boldsymbol{v}}_i$ and $\tilde{\boldsymbol{v}}_j$ are considered child of $\tilde{\boldsymbol{a}}_{i,j}$, and $\tilde{\boldsymbol{v}}_k$ belong to a child of $\tilde{\boldsymbol{a}}_{ijk}$.

\subsubsection{\textbf{Attribute-category Hierarchical Learning}}
Similarly, we model the attribute-category hierarchical structure between attribute prompts and text categories. We construct a series of triplets from the attribute prompts $\{\tilde{\boldsymbol{a}}_i, \tilde{\boldsymbol{a}}_j, \tilde{\boldsymbol{a}}_k\}$, where $\{\tilde{\boldsymbol{a}}_i, \tilde{\boldsymbol{a}}_j\}$ forms a positive pair, and $\{\tilde{\boldsymbol{a}}_i, \tilde{\boldsymbol{a}}_k\}$ denote negative pair. We employ the Eq. (\ref{psotive}) to identify the LCA $(\tilde{\boldsymbol{t}}_{ij})$ for the positive pair and Eq. (\ref{entire}) to obtain the LCA $(\tilde{\boldsymbol{t}}_{ijk})$ for the entire triplet respectively.
The regularization loss for the attribute-category  hierarchical structure is defined as:
\begin{equation}
\begin{split}
\mathcal{L}_{a-t}& =\mathrm{ReLU}[d_H(\tilde{\boldsymbol{a}}_i, \tilde{\boldsymbol{t}}_{ij})-d_H(\tilde{\boldsymbol{a}}_i, \tilde{\boldsymbol{t}}_{ijk})+\sigma]\\
& +\mathrm{ReLU}[d_H(\tilde{\boldsymbol{a}}_j,\tilde{\boldsymbol{t}}_{ij})-d_H(\tilde{\boldsymbol{a}}_j, \tilde{\boldsymbol{t}}_{ijk})+\sigma]\\
& +\mathrm{ReLU}[d_H(\tilde{\boldsymbol{a}}_k, \tilde{\boldsymbol{t}}_{ijk})-d_H(\tilde{\boldsymbol{a}}_k, \tilde{\boldsymbol{t}}_{ij})+\sigma],\\
\end{split}
\end{equation}
where $d_{H}(\cdot)$ and $\sigma$ are defined as the same as Eq. (\ref{av_reg}).


\subsection{Tuning and Inference}
During fine-tuning, similar to many other works~\cite{khattak2023maple,zhou2022conditional}, we utilize a contrastive loss to align visual and textual representations, which is formulated as follows:
\begin{equation}
\mathcal{L}_{ecls}=-\frac{1}{B}\sum_{i=1}^{B}\mathrm{log}\frac{\mathrm{exp}(\mathrm{cos}(\boldsymbol{v}_i,\hat{\mathbf{t}}_i)/\tau)}{\sum_{j=1}^{C}\mathrm{exp}(\mathrm{cos}(\boldsymbol{v}_i,\hat{\mathbf{t}}_j)/\tau)},
\end{equation}
where ${\boldsymbol{v}}$ represents a visual embedding, $\hat{\mathbf{t}}$ denotes the refined textual feature. 
$C$ denotes the number of classes and $B$ represents the number of image samples in each batch.
To leverage the complementary geometric properties of Euclidean and hyperbolic spaces, we further introduce an additional contrastive loss in hyperbolic space, defined as:
\begin{equation}
\mathcal{L}_{hcls}=-\frac{1}{B}\sum_{i=1}^{B}\mathrm{log}\frac{\mathrm{exp}(-d_H(\tilde{\boldsymbol{v}}_i,\tilde{\mathbf{t}}_i))}{\sum_{j=1}^{C}\mathrm{exp}(-d_H(\tilde{\boldsymbol{v}}_i,\tilde{\mathbf{t}}_j))},
\end{equation}
where ${\tilde{\boldsymbol{v}}}$ and $\tilde{\mathbf{t}}$ indicate the corresponding visual and textual embeddings in the hyperbolic space. The function $d_H(\cdot)$ represents the hyperbolic distance, which is defined in Eq. (\ref{distance}).   
Overall, the total loss for our LatHAdapter is denoted as:
\begin{equation}
\mathcal{L}=\mathcal{L}_{ecls}+\mathcal{L}_{hcls}+\mathcal{L}_{v-a}+\mathcal{L}_{a-t}.
\end{equation}

In inference, we integrate both Euclidean cosine similarity and hyperbolic distance to obtain the class probability. The probability for an image $\mathbf{X}^v$ is calculated as follows:
\begin{equation}
P(y=i|\mathbf{X^v})=\frac{\mathrm{exp}(\mathrm{cos}(\boldsymbol{v},\hat{\boldsymbol{t}}_i)/\tau-d_H(\tilde{\boldsymbol{v}},\tilde{\boldsymbol{t}}_i))}{\sum_{j=1}^{C}\mathrm{exp}(\mathrm{cos}(\boldsymbol{v},\hat{\boldsymbol{t}_j})/\tau-d_H(\tilde{\boldsymbol{v}},\tilde{\boldsymbol{t}}_j))},
\end{equation}
where $\boldsymbol{v}$ and $\tilde{\boldsymbol{v}}$ denote the visual features of image $\mathbf{X}^v$ in Euclidean and hyperbolic spaces, respectively.

\section{Comparison with Related Works} 
First,  LatHAdapter is significantly different from previous HyperCLIP~\cite{Open2025} and HypLoRA~\cite{yang2024hyperbolic}, which also employ hyperbolic learning in PEFT. 
HyperCLIP~\cite{Open2025} aims to capture pixel-level hierarchical structure of image data and proposes to adjust the hyperbolic radius of the textual embeddings to facilitate better alignment between text and image pixels. 
HypLoRA~\cite{yang2024hyperbolic} proposes a low-rank Lorentz transformation to adapt the model in the hyperbolic space.
Differently, our LatHAdapter aims to learn the latent semantic hierarchies from each batch of training data for downstream tasks and exploit them to provide fine-grained guidance for adapter learning. 
Second, our LatHAdapter is also different from some other hyperbolic-based multi-modal learning tasks, such as MERU~\cite{desai2023hyperbolic}, HyperLearner~\cite{kong2024hyperbolic}. These methods leverage the entailment cone to model the intrinsic hierarchical structure of existing multi-modal data. Differently, our LatHAdapter introduces attribute prompts as intermediate semantic representations between categories and images, and models the inherent hierarchical \emph{triplet} representation (category $\rightarrow$ attribute $\rightarrow$ image), thereby enhancing fine-grained textual representations for both known and unknown classes. 
Finally, LatHAdapter is obviously different from ATPrompt~\cite{li2025atprompt}, which achieves textual prompt learning with embedded attributes. 
ATPrompt~\cite{li2025atprompt} proposes to first determine some explicit attributes and then employ them for textual prompt learning. 
In contrast, LatHAdapter treats the `attributes' as the learnable prompts and obtains them adaptively with no annotation.
Also, it employs hyperbolic learning to further exploit the inherent semantic hierarchies among categories, attributes, and image samples.

\section{Experiments}
\label{experiment}
\subsection{Experimental Setup}
\begin{table*}[!h]
 \caption{Base-to-new generalization performance of three representative fine-tuning methods w/ or w/o our LatHAdapter on 11 datasets. By modeling semantic hierarchies in hyperbolic space, LatHAdapter consistently achieves the best performance on both base and new classes.
}\label{base_new}
    \centering
    \renewcommand{\arraystretch}{0.8}
\begin{tabular}{c|ccc|ccc|ccc|ccc} 
\toprule
\multirow{2.4}{*}{Method} & \multicolumn{3}{c|}{\textbf{Avg. over 11 datasets}} & \multicolumn{3}{c|}{ImageNet}   & \multicolumn{3}{c|}{Caltech101} & \multicolumn{3}{c}{OxfordPets}    \\ 
\cmidrule(lr){2-4}\cmidrule(lr){5-7}\cmidrule(lr){8-10}\cmidrule(lr){11-13}
                        & Base  & New   & H                            & Base  & New   & H              & Base  & New   & H              & Base  & New   & H                 \\ 
\midrule
CLIP \textsubscript{\tiny [ICML2021]} &69.34 &74.22 &71.70 &72.43 &68.14 &70.22 &96.84 &94.00 &95.40 &91.17 &97.26 &94.12\\
COMMA \textsubscript{\tiny [AAAI2024]} &82.42 &75.87 &79.04 &76.04 &70.89 &73.86 &97.94 &94.56 &96.50 &95.62 &97.84 &96.72\\
MMA \textsubscript{\tiny [CVPR2024]} &83.20 &76.80 &79.87 &77.31 &71.00 &74.02 &98.40 &94.00 &96.15 &95.40 &98.07 &96.72\\
TCP \textsubscript{\tiny [CVPR2024]}  &84.13 &75.36 &79.51 &77.27 &69.87 &73.38 &98.23 &94.67 &96.42 &94.67 &97.20 &95.92 \\
PromptSRC+DePT \textsubscript{\tiny [CVPR2024]} &84.08 &75.03 &79.29 &77.91 &70.77 &74.17 &98.37 &94.14 &96.21 &94.83 &97.21 &96.00 \\
PromptSRC+DPC \textsubscript{\tiny [CVPR2025]}    &86.10 &74.78 &80.04 &78.48 &70.72 &74.40 &98.90 &94.21 &96.50 &96.13 &97.30 &96.71 \\
PromptSRC+ATP \textsubscript{\tiny [ICCV2025]}    &84.30 &76.45 &80.18 &77.69 &70.83 &74.10 &98.23 &94.91 &96.54 &95.64 &97.43 &96.53 \\
\midrule
MaPLe \textsubscript{\tiny[CVPR2023]}                  & 81.22 & 75.08 & 78.03                        & 76.89 & 70.48 & 73.54          & 97.91 & 94.39 & 96.12          & 95.44 & 98.02 & 96.71             \\
\cellcolor{cyan!15}\textbf{+LatHAdapter} & \cellcolor{cyan!15}\textbf{82.42} & \cellcolor{cyan!15}\textbf{76.13} & \cellcolor{cyan!15}\textbf{79.15}                        & \textbf{76.99} & \textbf{70.49} & \textbf{73.60}          & \textbf{98.15} & \textbf{94.87} & \textbf{96.48}          & \textbf{96.07} & \textbf{98.04} & \textbf{97.05}             \\
$\triangle$& \textcolor{blue}{(+1.20)} & \textcolor{blue}{(+1.05)} & \textcolor{blue}{(+1.12)} & \textcolor{blue}{(+0.10)} & \textcolor{blue}{(+0.01)} & \textcolor{blue}{(+0.06)} & \textcolor{blue}{(+0.24)} & \textcolor{blue}{(+0.48)} & \textcolor{blue}{(+0.36)} & \textcolor{blue}{(+0.63)} & \textcolor{blue}{(+0.02)} & \textcolor{blue}{(+0.34)}\\
\midrule
PromptSRC \textsubscript{\tiny [ICCV2023]}              & 84.14 & 75.65 & 79.67                        & 77.67 & 70.22 & 73.82          & 98.03 & 93.83 & 95.88          & 95.36 & \textbf{97.30} & 96.32             \\
\cellcolor{cyan!15}\textbf{+LatHAdapter} & \cellcolor{cyan!15}\textbf{84.68} & \cellcolor{cyan!15}\textbf{76.52} & \cellcolor{cyan!15}\textbf{80.40}                        & \textbf{77.84} & \textbf{70.58} & \textbf{74.03}          & \textbf{98.32} & \textbf{94.43} & \textbf{96.34}          & \textbf{97.37} & 97.22 & \textbf{97.29}             \\ 
$\triangle$& \textcolor{blue}{(+0.54)} & \textcolor{blue}{(+0.87)} & \textcolor{blue}{(+0.73)} & \textcolor{blue}{(+0.17)} & \textcolor{blue}{(+0.36)} & \textcolor{blue}{(+0.21)} & \textcolor{blue}{(+0.29)} & \textcolor{blue}{(+0.60)} & \textcolor{blue}{(+0.46)} & \textcolor{blue}{(+2.01)} & \textcolor{blue}{(-0.08)} & \textcolor{blue}{(+0.97)} \\
\midrule
PromptKD \textsubscript{\tiny [CVPR2024]}                   & 84.15 & 78.98 & 81.48                        & \textbf{77.20} & 70.90 & 73.92       & 98.50 & 96.90 & 97.69          & 94.50& 96.80 & 95.64             \\
\cellcolor{cyan!15}\textbf{+LatHAdapter} & \cellcolor{cyan!15}\textbf{84.96} & \cellcolor{cyan!15}\textbf{79.92} & \cellcolor{cyan!15}\textbf{82.37}                        & \textbf{77.20} & \textbf{71.00} & \textbf{73.97}          & \textbf{98.60} & \textbf{97.10} & \textbf{97.84}          & \textbf{95.00} & \textbf{97.80} & \textbf{96.38}             \\
$\triangle$& \textcolor{blue}{(+0.81)} & \textcolor{blue}{(+0.94)} & \textcolor{blue}{(+0.89)} & \textcolor{blue}{(+0.00)} & \textcolor{blue}{(+0.10)} & \textcolor{blue}{(+0.05)} & \textcolor{blue}{(+0.10)} & \textcolor{blue}{(+0.20)} & \textcolor{blue}{(+0.15)} & \textcolor{blue}{(+0.50)} & \textcolor{blue}{(+1.00)} & \textcolor{blue}{(+0.74)}\\
\midrule\midrule
\multirow{2.4}{*}{Method} & \multicolumn{3}{c|}{StanfordCars}             & \multicolumn{3}{c|}{Flowers102} & \multicolumn{3}{c|}{Food101}    & \multicolumn{3}{c}{FGVCAircraft}  \\ 
\cmidrule(lr){2-4}\cmidrule(lr){5-7}\cmidrule(lr){8-10}\cmidrule(lr){11-13}
                        & Base  & New   & H                            & Base  & New   & H              & Base  & New   & H              & Base  & New   & H                 \\ 
\midrule
CLIP \textsubscript{\tiny [ICML2021]} &63.37 &74.89 &68.65 &72.08 &77.80 &74.83 &90.10 &91.22 &90.66 &27.19 &36.29 &31.09\\
COMMA \textsubscript{\tiny [AAAI2024]} &73.48 &74.91 &73.96 &94.86 &75.13 &83.88 &90.42 &92.74 &91.84 &36.47 &34.23 &35.84\\
MMA \textsubscript{\tiny [CVPR2024]} &78.50 &73.10 &75.70 &97.77 &75.93 &85.48 &90.13 &91.30 &90.71 &40.57 &36.33 &38.33\\
TCP \textsubscript{\tiny [CVPR2024]}  &80.80 &74.13 &77.32 &97.73 &75.57 &85.23 &90.57 &91.37 &90.97 &41.97 &34.43 &37.83 \\
PromptSRC+DePT \textsubscript{\tiny [CVPR2024]} &78.26 &74.73 &76.46 &97.44 &74.89 &84.69 &90.61 &91.63 &91.12 &41.18 &35.63 &38.20 \\
PromptSRC+DPC \textsubscript{\tiny [CVPR2025]}    &82.28 &74.98 &78.46 &97.44 &73.19 &83.59 &91.40 &91.58 &91.49 &46.74 &35.33 &40.24 \\
PromptSRC+ATP \textsubscript{\tiny [ICCV2025]}    &79.25 &74.95 &77.04 &97.82 &77.02 &86.18 &90.77 &91.78 &91.27 &42.47 &37.01 &39.55 \\
\midrule
MaPLe \textsubscript{\tiny[CVPR2023]}                   & 71.83 & \textbf{74.27} & 72.03                        & 95.28 & 73.55 & 83.02          & 90.77 & 91.88 & 91.32          & 36.16 & 34.39 & 35.25             \\
\textbf{+LatHAdapter}                    & \textbf{73.06} & 73.93 & \textbf{73.49}                        & \textbf{96.04} & \textbf{74.44} & \textbf{83.87}          & \textbf{90.85} & \textbf{91.90} & \textbf{91.37}          & \textbf{37.49} & \textbf{35.99} & \textbf{36.73}             \\ 
$\triangle$& \textcolor{blue}{(+1.25)} & \textcolor{blue}{(-0.34)} & \textcolor{blue}{(+1.46)} & \textcolor{blue}{(+0.76)} & \textcolor{blue}{(+0.89)} & \textcolor{blue}{(+0.85)} & \textcolor{blue}{(+0.08)} & \textcolor{blue}{(+0.02)} & \textcolor{blue}{(+0.05)} & \textcolor{blue}{(+1.33)} & \textcolor{blue}{(+1.60)} & \textcolor{blue}{(+1.48)}\\
\midrule
PromptSRC \textsubscript{\tiny [ICCV2023]}                & 78.27 & \textbf{75.27} & \textbf{76.78}                        & \textbf{97.87} & 76.63 & 85.96          & 90.80 & 91.47 & 91.33          & \textbf{43.10} & 37.20 & 39.93             \\
\textbf{+LatHAdapter}                    & \textbf{78.94} & 74.35 & 76.58                        & 96.9 & \textbf{78.27} & \textbf{86.59}          & \textbf{91.34}& \textbf{93.17}& \textbf{92.25}          & 42.17 & \textbf{38.50} & \textbf{40.25}             \\ 
$\triangle$& \textcolor{blue}{(+0.67)} & \textcolor{blue}{(-0.92)} & \textcolor{blue}{(-0.20)} & \textcolor{blue}{(-0.97)} & \textcolor{blue}{(+1.64)} & \textcolor{blue}{(+0.63)} & \textcolor{blue}{(+0.54)} & \textcolor{blue}{(+1.70)} & \textcolor{blue}{(+0.92)} & \textcolor{blue}{(-0.93)} & \textcolor{blue}{(+1.30)} & \textcolor{blue}{(+0.32)}\\
\midrule
PromptKD \textsubscript{\tiny [CVPR2024]}                & \textbf{80.60} & 82.40 & \textbf{81.49}                        & 98.80 & 82.00 & 89.62        & \textbf{89.50} & \textbf{91.70} & \textbf{90.59}          & 45.70 & \textbf{44.10} & \textbf{44.88}             \\
\textbf{+LatHAdapter}                    & 80.30 & \textbf{82.40} & 81.34                       & \textbf{98.90} & \textbf{82.80} & \textbf{90.14}          & 89.50 & 91.50 & 90.49          & \textbf{46.20} & 40.90 & 43.39            \\ 
$\triangle$& \textcolor{blue}{(-0.30)} & \textcolor{blue}{(+0.00)} & \textcolor{blue}{(-0.15)} & \textcolor{blue}{(+0.10)} & \textcolor{blue}{(+0.80)} & \textcolor{blue}{(+0.52)} & \textcolor{blue}{(+0.54)} & \textcolor{blue}{(+1.70)} & \textcolor{blue}{(+0.92)} & \textcolor{blue}{(-0.93)} & \textcolor{blue}{(+1.30)} & \textcolor{blue}{(+0.32)}\\
\midrule\midrule
\multirow{2.4}{*}{Method} & \multicolumn{3}{c|}{SUN397}                   & \multicolumn{3}{c|}{DTD}        & \multicolumn{3}{c|}{EuroSAT}    & \multicolumn{3}{c}{UCF101}        \\ 
\cmidrule(lr){2-4}\cmidrule(lr){5-7}\cmidrule(lr){8-10}\cmidrule(lr){11-13}
                        & Base  & New   & H                            & Base  & New   & H              & Base  & New   & H              & Base  & New   & H                 \\ 
\midrule
CLIP \textsubscript{\tiny [ICML2021]} &69.36 &75.35 &72.23 &53.24 &59.90 &56.37 &56.48 &64.05 &60.03 &70.53 &77.50 &73.85\\
COMMA \textsubscript{\tiny [AAAI2024]} &80.94 &79.32 &80.86 &81.04 &58.62 &68.32 &93.56 &74.26 &83.42 &84.06 &80.56 &81.84\\
MMA \textsubscript{\tiny [CVPR2024]}&82.27 &78.57 &80.38 &83.20 &65.63 &73.38 &85.46 &82.34 &83.87 &86.23 &80.03 &82.20\\
TCP \textsubscript{\tiny [CVPR2024]}  &82.63 &78.20 &80.35 &82.77 &58.07 &68.25 &91.63 &74.73 &82.32 &87.13 &80.77 &83.83 \\
PromptSRC+DePT \textsubscript{\tiny [CVPR2024]}&82.60 &78.82 &80.67 &83.64 &59.18 &69.32 &94.46 &71.01 &81.07 &85.54 &77.29 &81.20 \\
PromptSRC+DPC \textsubscript{\tiny [CVPR2025]}   &83.63 &78.08 &80.76 &86.88 &54.31 &66.84 &96.25 &74.73 &84.13 &88.99 &78.13 &83.21 \\
PromptSRC+ATP \textsubscript{\tiny [ICCV2025]}    &82.73 &78.64 &80.63 &83.22 &62.68 &71.50 &92.29 &76.42 &83.61 &87.15 &79.23 &83.00 \\
\midrule
MaPLe \textsubscript{\tiny[CVPR2023]}                  & \textbf{81.03} & 78.21 & 79.59                        & 77.55 & 59.14 & 67.10          & 88.08 & 76.50 & 81.88          & 82.54 & 75.01 & 78.59             \\
\textbf{+LatHAdapter}                    & 80.95 & \textbf{78.74} & \textbf{79.83}                        & \textbf{79.63} & \textbf{61.96} & \textbf{69.69}          & \textbf{93.66} & \textbf{77.69} & \textbf{84.93}          & \textbf{83.73} & \textbf{79.38} & \textbf{81.50}             \\ 
$\triangle$ & \textcolor{blue}{(-0.35)} & \textcolor{blue}{(+0.53)} & \textcolor{blue}{(+0.24)} & \textcolor{blue}{(+2.08)} & \textcolor{blue}{(+2.82)} & \textcolor{blue}{(+2.59)} & \textcolor{blue}{(+5.58)} & \textcolor{blue}{(+1.19)} & \textcolor{blue}{(+3.05)} & \textcolor{blue}{(+1.19)} & \textcolor{blue}{(+4.37)} & \textcolor{blue}{(+2.91)}\\
\midrule
PromptSRC \textsubscript{\tiny [ICCV2023]}                & 82.47 & 78.73 & 80.56                        & \textbf{82.90} & \textbf{62.67} & \textbf{71.38}          & \textbf{92.83} & 70.06 & 79.85          & 86.17 & \textbf{78.67} & 82.25          \\
\textbf{+LatHAdapter}                    & \textbf{82.65} & \textbf{78.88} & \textbf{80.72}                        & 82.29 & 62.32 & 70.93          & 92.79 & \textbf{76.52} & \textbf{83.87}          & \textbf{90.85} & 77.5 & \textbf{83.65}             \\ 
$\triangle$ & \textcolor{blue}{(+0.18)} & \textcolor{blue}{(+0.15)} & \textcolor{blue}{(+0.16)} & \textcolor{blue}{(-0.61)} & \textcolor{blue}{(-0.35)} & \textcolor{blue}{(-0.45)} & \textcolor{blue}{(-0.04)} & \textcolor{blue}{(+6.46)} & \textcolor{blue}{(+4.02)} & \textcolor{blue}{(+4.68)} & \textcolor{blue}{(-1.17)} & \textcolor{blue}{(+1.40)}\\
\midrule
PromptKD \textsubscript{\tiny [CVPR2024]}                  & \textbf{83.20} & 80.30 & 81.72                        & 82.40 & 69.10 & 75.16          & 87.20 & 74.20 & 80.17          & \textbf{88.10} & 80.40 & 84.07             \\
\textbf{+LatHAdapter}                    & 82.80 & \textbf{81.00} & \textbf{81.89}                        & \textbf{82.90} & \textbf{72.10} & \textbf{77.12}          & \textbf{95.40} & \textbf{81.40} & \textbf{87.85}          & 87.80 & \textbf{81.10} & \textbf{84.32}             \\
$\triangle$ & \textcolor{blue}{(-0.40)} & \textcolor{blue}{(+0.70)} & \textcolor{blue}{(+0.17)} & \textcolor{blue}{(+0.50)} & \textcolor{blue}{(+3.00)} & \textcolor{blue}{(+1.96)} & \textcolor{blue}{(+8.20)} & \textcolor{blue}{(+7.20)} & \textcolor{blue}{(+7.68)} & \textcolor{blue}{(-0.30)} & \textcolor{blue}{(+0.70)} & \textcolor{blue}{(+0.25)}\\
\bottomrule
\end{tabular}
\end{table*}
\subsubsection{Datasets}
We follow previous works~\cite{khattak2023maple,seputis2024multi,khattak2023self} and evaluate our method on four few-shot tasks: base-to-new generalization, few-shot image classification, cross-dataset transfer, and domain generalization. For the domain generalization task, we perform experiments on four datasets, including ImageNet-V2~\cite{recht2019imagenet}, ImageNet-Sketch~\cite{wang2019learning}, ImageNet-A~\cite{hendrycks2021natural}, and ImageNet-R~\cite{hendrycks2021many}. For base-to-new generalization, cross-dataset transfer, and few-shot image classification tasks, we evaluate the proposed LatHAdapter on 11 datasets, including ImageNet~\cite{deng2009imagenet}, Caltech101~\cite{fei2004learning}, OxfordPets~\cite{parkhi2012cats}, StanfordCars~\cite{krause20133d}, Flowers102~\cite{nilsback2008automated}, Food101~\cite{bossard2014food}, FGVCAircraft~\cite{maji2013fine}, SUN397~\cite{xiao2010sun}, DTD~\cite{cimpoi2014describing}, EuroSAT~\cite{helber2019eurosat}, and UCF101~\cite{soomro2012ucf101}. These datasets cover diverse scenarios, such as fine-grained classification, action recognition, texture classification, and remote sensing classification.

\subsubsection{Implementation details}
To verify the effectiveness of the proposed LatHAdapter, we integrate it into three representative Parameter-Efficient Fine-Tuning (PEFT) frameworks: MaPLe~\cite{khattak2023maple}, PromptSRC~\cite{khattak2023self}, and PromptKD~\cite{li2024promptkd}. MaPLe explores the layer-wise interaction between text and image encoders. PromptSRC utilizes separate visual and text prompts for fine-tuning. PromptKD adopts ViT-L/14 as the teacher model to guide student adaptation via knowledge distillation. We follow the default hyperparameter settings provided in each baseline. For a fair comparison, ViT-B/16 CLIP architecture is employed as the backbone for all methods, and the batch size is fixed to 4 for all experiments. Specifically, we fine-tune the VLMs using a few-shot set of image–text pairs, without using any unlabeled images. The number of attribute prompts is set based on the characteristics of each dataset. For hyperbolic modeling, we set the curvature $c=0.1$ and $\sigma=0.1$, which follows prior work~\cite{kim2023hier}. The balance factor $\beta$ is set to 0.1 in all experiments to control the trade-off between preserving the original CLIP knowledge and adapting to the downstream task. We set the number of nearest neighbors $k=3$ to obtain positive and negative pairs.

\subsection{Comparison Results}
\subsubsection{Base-to-new generalization}
For the base-to-new task, each dataset is split into base classes and new classes. Vision-language models (VLMs) are fine-tuned in base classes via a $16$-shot setting and evaluated in both base and new classes. To assess the effectiveness of the proposed plug-and-play LatHAdapter, we integrate it into three representative fine-tuning methods: MaPLe~\cite{khattak2023maple}, PromptSRC~\cite{khattak2023self}, and PromptKD~\cite{li2024promptkd}. Furthermore, we compare these methods with other state-of-the-art (SOTA), including CLIP~\cite{radford2021learning}, COMMA~\cite{hu2024comma}, MMA~\cite{seputis2024multi},  and plug-and-play approaches such as PromptSRC+DePT~\cite{zhang2024dept}, TCP~\cite{yao2024tcp}, PromptSRC+DPC~\cite{li2025dpc}, and PromptSRC+ATP~\cite{li2025atprompt}. We report accuracy on both base (Base) and new classes (New), while the harmonic mean (H) provides a comprehensive metric to evaluate the performance between them. As shown in Tab.~\ref{base_new}, integrating LatHAdapter with three representative methods consistently improves the average performance on 11 datasets. We observe H gains of $1.12\%$, $0.73\%$, and $0.89\%$ for MaPLe, PromptSRC, and PromptKD, respectively. Specifically, MaPLe with LatHAdapter improves accuracy on all 11 datasets, while PromptSRC and PromptKD achieve gains on 9/11 and 8/11 datasets, respectively. Compared to existing plug-and-play methods, the proposed LatHAdapter also achieves the best performance. These results demonstrate that the proposed LatHAdapter effectively models the latent semantic hierarchy for the downstream training data and learns transferable and discriminative attribute prompts, thereby significantly improving generalization on both base and new classes.

\subsubsection{Few-shot image classification}
\begin{figure*}[!h]
  \centering
  \includegraphics[width=1\linewidth]{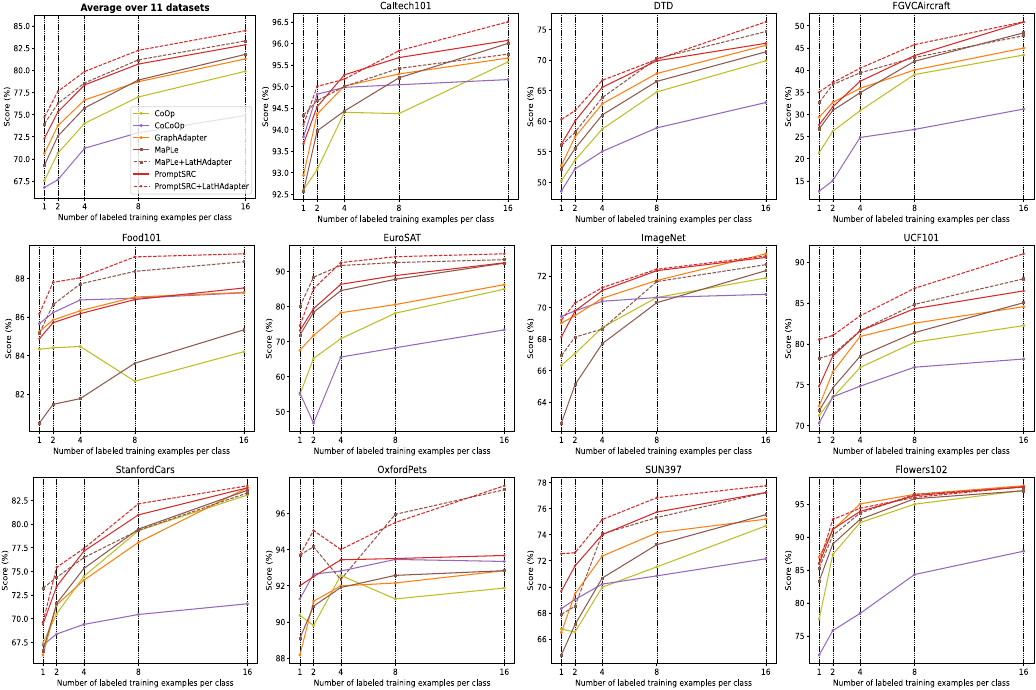}
\caption{The performance of two representative baseline methods w/ or w/o our LatHAdapter in few-shot image classification tasks, including 1-/2-/4-/8-/16-shots on 11 datasets. All methods are trained on ViT-B/16 CLIP backbone.} 
\label{few_shot}
\end{figure*}

To validate the effectiveness of the proposed LatHAdapter in learning task-specific knowledge, we perform experiments on few-shot image classification tasks. We integrate LatHAdapter into MaPLe~\cite{khattak2023maple} and PromptSRC~\cite{khattak2023self}, which is denoted as ``MaPLe+LatHAdapter'' and ``PromptSRC+LatHAdapter'' in Fig.~\ref{few_shot}. Furthermore, we compare our methods with several state-of-the-art few-shot image classification methods, such as CoOp~\cite{zhou2022learning}, CoCoOp~\cite{zhou2022conditional}, Graph-adapter~\cite{li2024graphadapter}, MaPLe~\cite{khattak2023maple}, and PromptSRC~\cite{khattak2023self}. All methods employ ViT-B/16 CLIP as a backbone for a fair comparison. By incorporating LatHAdapter with PromptSRC, we achieve consistent performance improvements across 11 datasets. The average accuracy increases by 2.39\%, 2.32\%, 1.51\%, 1.55\%, and 1.60\% for the 1-shot, 2-shot, 4-shot, 8-shot, and 16-shot settings, respectively. Meanwhile, ``MaPLe+LatHAdapter'' shows consistent average performance gains of 4.68\%, 3.66\%, 2.82\%, 2.28\%, and 1.52\% for the 1-shot, 2-shot, 4-shot, 8-shot, and 16-shot settings, respectively. Significant improvements are observed in the 1-shot and 2-shot settings, highlighting the effectiveness of LatHAdapter in modeling latent semantic hierarchies to enhance few-shot learning performance, especially under limited labeled data. The numerical results corresponding to Fig.\ref{few_shot} are presented in Tab.\ref{tab_few_shot}.

\subsubsection{Cross-dataset and domain generalization}
In cross-dataset transfer tasks, we fine-tune the VLMs on ImageNet with a 16-shot setting and evaluate their zero-shot generation ability on other datasets. Tab.~\ref{cross} reports the performance comparison between MaPLe, PromptSRC (with and without LatHAdapter), and other state-of-the-art (SOTA) methods, including  CoCoOp~\cite{zhou2022conditional}, MMA~\cite{seputis2024multi}, and TCP~\cite{yao2024tcp}. The MaPLe+LatHAdapter and PromptSRC+LatHAdapter achieve competitive results across 11 datasets, with average performance improvements of +0.43\% and +0.59\%, respectively. To assess the domain generalization capability of LatHAdapter, we conduct experiments on four out-of-distribution datasets: ImageNet-V2, ImageNet-Sketch, ImageNet-A, and ImageNet-R. We directly evaluate the model fine-tuned on the ImageNet dataset, and the comparison results are presented in Tab.~\ref{tab2}. MaPLe+LatHAdapter and PromptSRC+LatHAdapter obtain an average accuracy of 60.55\% and 60.71\% across four datasets, consistently outperforming their baseline method. These results demonstrate the effectiveness of the proposed plug-and-play LatHAdapter in preserving both zero-shot and domain generalization performance.

\begin{table*}[!h]
\caption{Cross-dataset transfer experiments of two baselines w/ or w/o our LatHAdapter on 11 datasets.}
\centering
\renewcommand{\arraystretch}{0.8}
\setlength\tabcolsep{2pt}
\begin{tabular}{c|c|c|cccccccccc} 
\toprule
\multirow{3.5}{*}{Method} & \textbf{Source}   & \multicolumn{10}{c}{\textbf{Target}}\\ 
\cmidrule(lr){2-2}\cmidrule(lr){3-13}
                        & ImageNet & Avg & Caltech101 & OxfordPets&StanfordCars& Flowers102&  Food101&  FGVCAircraft&  SUN397&  DTD&  EuroSAT&  UCF101           \\ 
                    \midrule
CoCoOp \textsubscript{\tiny [CVPR2022]} &71.02 &65.74  &94.43 &90.14 &65.32 &71.88 &86.06 &22.94 &67.36 &45.73 &45.37 &68.21  \\ 
MMA \textsubscript{\tiny [CVPR2024]}   &71.00 &66.61 &93.80 &90.30 &66.13 &72.07 &86.12 &25.33 &68.17 &46.57 &49.24 &68.32 \\
TCP \textsubscript{\tiny [CVPR2024]} &71.40 &66.29 &93.97 & 91.25& 64.69 & 71.21 & 86.69 & 23.45 & 67.15 & 44.35 & 51.45 & 68.73 \\
\midrule
MaPLe \textsubscript{\tiny [CVPR2023]}   & 70.72    & 66.30   &\textbf{93.53} & \textbf{90.49} & 65.57 & \textbf{72.23} & \textbf{86.20} & 24.74 & 67.01 & 46.49 & 48.06 & 68.69
\\
\rowcolor{gray!20}
\textbf{+LatHAdapter} & 70.61    & \textbf{66.73}    & 93.51 & 90.13 & \textbf{65.86} & 70.93 & 86.00 & \textbf{25.35} & \textbf{67.02} & \textbf{46.81} & \textbf{52.63} & \textbf{69.07}\\
\midrule
PromptSRC \textsubscript{\tiny [ICCV2023]}  & 71.27    & 65.81    & 93.60 & \textbf{90.25} & \textbf{65.70} & 70.25 & 86.15 & 23.90 & \textbf{67.10} & 46.87 & 45.50 & \textbf{68.75}
\\
\rowcolor{gray!20}
\textbf{+LatHAdapter} & 71.77    & \textbf{66.40}   & \textbf{93.87} & 89.92 & 64.86 & \textbf{70.65} & \textbf{86.22} & \textbf{24.30} & 66.79 & \textbf{48.46} & \textbf{51.04} & 67.88\\              
\bottomrule
\end{tabular}
\label{cross}
\end{table*}

\begin{table}[!htbp]
\caption{Cross-domain generalization experiments of two baselines w/ or w/o LatHAdapter.}
    \centering
    \renewcommand{\arraystretch}{0.8}
\setlength\tabcolsep{2pt}
\begin{tabular}{c|c|c|cccc} 
\toprule
\multirow{3.5}{*}{Method} & \textbf{Source}   & \multicolumn{5}{c}{\textbf{Target}}\\ 
\cmidrule(lr){2-2}\cmidrule(lr){3-7}
                        & ImageNet & Avg & -V2 & -Sketch & -A & -R           \\
\midrule
CoCoOp \textsubscript{\tiny [CVPR2022]} &71.02 &59.91 &64.07 &48.75 &50.63 &76.18 \\
MMA \textsubscript{\tiny [CVPR2024]} &71.00 &60.47 &64.33 &49.13 &51.12 &77.32 \\
\midrule
MaPLe\textsubscript{\tiny [CVPR2023]}   & 70.72    & 60.27   &\textbf{64.07} & 49.15 & 50.90 & 76.98 
\\
\rowcolor{gray!20}
\textbf{+LatHAdapter} & 70.61    & \textbf{60.55}    & 64.02 & \textbf{49.28} & \textbf{51.35} & \textbf{77.53} \\
\midrule
PromptSRC\textsubscript{\tiny [ICCV2023]}   & 71.27    & 60.65    & 64.35 & \textbf{49.55} & 50.90 & \textbf{77.80}
\\
\rowcolor{gray!20}
\textbf{+LatHAdapter} & 71.77    & \textbf{60.71}   & \textbf{64.89} & 49.39 & \textbf{51.12} & 77.45 \\              
\bottomrule
\end{tabular}
\label{tab2}
\end{table}


\begin{table}[!h]
\caption{Ablation study of the Hyperbolic Hierarchical Learning (HHL) in the base-to-new task. Experiments are conducted on the PromptSRC baseline.}\label{HiAL}
    \centering
    \renewcommand{\arraystretch}{0.8}
\begin{tabular}{c|ccc} 
\toprule
\multirow{2.4}{*}{Method} & \multicolumn{3}{c}{\textbf{Avg. over 11 datasets}}   \\ 
\cmidrule(lr){2-4}
                & Base  & New   & H                          \\ 
\midrule
PromptSRC               & 84.14 & 75.65 & 79.67      \\
\midrule
PromptSRC+ATR          & 83.67 & 75.33 & 79.28   \\
PromptSRC+LatHAdapter (ATR+HHL)     & 84.68 & 76.52 & 80.40   \\
\bottomrule
\end{tabular}
\end{table}

\begin{table*}[!h]
  \centering
  \caption{Numerical comparison results for few-shot learning across 1-/2-/4-/8-/16-shot settings on 11 datasets.}
   \resizebox{1\textwidth}{!}{
    \begin{tabular}{c|c|ccccccccccc|c}
    \toprule
    {Method} & Setting & Caltech101 & DTD   & EuroSAT & FGVCAircraft & Flowers102 & Food101 & ImageNet & OxfordPets & StanfordCars & SUN397 & UCF101 & AVG  \\
    \midrule
    {CoOp } & \multirow{8}[1]{*}{1} & 92.60  & 50.23 & 54.93 & 21.37 & 77.53 & 84.33 & 66.33 & 90.37 & 67.43 & 66.77 & 71.23 & 67.56  \\
    {CoCoOp} &       & 93.83 & 48.54 & 55.33 & 12.68 & 72.08 & 85.65 & 69.43 & 91.27 & 67.22 & 68.33 & 70.30  & 66.79  \\
    {GraphAdapter} & &92.94	&52.72	&67.54	&29.37	&86.93	&85.24	&69.01	&88.17	&66.15	&66.50	&72.40 &70.63\\
    {MaPLe} &       & 92.57 & 52.13 & 71.80  & 26.73 & 83.30  & 80.50  & 62.67 & 89.10  & 66.60  & 64.77 & 71.83 & 69.27  \\
    {+LatHAdapter } &       & 94.33 & 56.03 & 80.00    & 32.67 & 85.29 & 85.15 & 66.94 & 93.69 & 73.22 & 67.89 & 78.22 & 73.95  \\
          $\triangle$ &    & \textcolor{blue}{(+1.76)} &\textcolor{blue}{(+3.90)} &\textcolor{blue}{(+8.20)} &\textcolor{blue}{(+5.94)} &\textcolor{blue}{(+1.99)} &\textcolor{blue}{(+4.65)} &\textcolor{blue}{(+4.27)} &\textcolor{blue}{(+4.59)} &\textcolor{blue}{(+6.62)} &\textcolor{blue}{(+3.12)} &\textcolor{blue}{(+6.39)} &\textcolor{blue}{(+4.68) } \\
    {PromptSRC} &       & 93.67 & 56.23 & 73.13 & 27.67 & 85.93 & 84.87 & 68.13 & 92.00  & 69.40  & 69.67 & 74.80  & 72.32  \\
    {+LatHAdapter } &       & 94.12 & 60.28 & 74.52 & 35.00  & 86.07 & 86.15 & 69.21 & 93.69 & 69.63 & 72.54 & 80.53 & 74.70  \\
          $\triangle$ &       &\textcolor{blue}{(+0.45)} &\textcolor{blue}{(+4.05)} &\textcolor{blue}{(+1.39)} &\textcolor{blue}{(+7.33)} &\textcolor{blue}{(+0.14)} &\textcolor{blue}{(+1.28)} &\textcolor{blue}{(+1.08)} &\textcolor{blue}{(+1.69)} &\textcolor{blue}{(+0.23)} &\textcolor{blue}{(+2.87)} &\textcolor{blue}{(+5.73)} &\textcolor{blue}{(+2.39) } \\
    \midrule
    {CoOp } & \multirow{8}[2]{*}{2} & 93.07 & 53.60  & 65.17 & 26.20  & 87.33 & 84.40  & 67.07 & 89.80  & 70.50  & 66.53 & 73.43 & 70.65   \\
    {CoCoOp} &       & 94.82 & 52.17 & 46.74 & 15.06 & 75.79 & 86.22 & 69.78 & 92.64 & 68.37 & 69.03 & 73.51 & 67.65  \\
    {GraphAdapter} & & 94.36	&57.39	&71.74	&32.70	&91.07	&85.84	&69.48	&91.14	&71.45	&69.45	&76.55 &73.74\\
    {MaPLe} &       & 93.97 & 55.50  & 78.30  & 30.90  & 88.93 & 81.47 & 65.10  & 90.87 & 71.60  & 67.10  & 74.60  & 72.58  \\
    {+LatHAdapter } &       & 94.67 & 58.15 & 88.33 & 36.75 & 90.20  & 86.63 & 68.12 & 94.15 & 74.36 & 68.51 & 78.71 & 76.23  \\
         $\triangle$   &       &\textcolor{blue}{(+0.70)} &\textcolor{blue}{(+2.65)} &\textcolor{blue}{(+10.03)} &\textcolor{blue}{(+5.85)} &\textcolor{blue}{(+1.27)} &\textcolor{blue}{(+5.16)} &\textcolor{blue}{(+3.02)} &\textcolor{blue}{(+3.28)} &\textcolor{blue}{(+2.76)} &\textcolor{blue}{(+1.41)} &\textcolor{blue}{(+4.11)} &\textcolor{blue}{(+3.66)} \\
    {PromptSRC} &       & 94.53 & 59.97 & 79.37 & 31.70  & 91.17 & 85.70  & 69.77 & 92.50  & 73.40  & 71.60  & 78.50  & 75.29  \\
    {+LatHAdapter } &       & 95.00  & 61.70  & 85.00    & 37.17 & 92.65 & 87.79 & 70.27 & 95.05 & 75.42 & 72.61 & 81.02 & 77.61  \\
          $\triangle$  &       &\textcolor{blue}{(+0.47)} &\textcolor{blue}{(+1.73)} &\textcolor{blue}{(+5.63)} &\textcolor{blue}{(+5.47)} &\textcolor{blue}{(+1.48)} &\textcolor{blue}{(+2.09)} &\textcolor{blue}{(+0.50)} &\textcolor{blue}{(+2.55)} &\textcolor{blue}{(+2.02)} &\textcolor{blue}{(+1.01)} &\textcolor{blue}{(+2.52)} &\textcolor{blue}{(+2.32) } \\
    \midrule
    {CoOp } & \multirow{8}[2]{*}{4} & 94.40  & 58.70  & 70.80  & 30.83 & 92.17 & 84.47 & 68.73 & 92.57 & 74.47 & 69.97 & 77.10  & 74.02   \\
    {CoCoOp} &       & 94.98 & 55.04 & 65.56 & 24.79 & 78.40  & 86.88 & 70.39 & 92.81 & 69.39 & 70.21 & 74.82 & 71.21  \\
    {GraphAdapter} & & 95.01	&62.83	&78.20	&35.88	&95.01	&86.32	&70.58	&91.99	&74.07	&72.36	&80.89 &76.65\\
    {MaPLe} &       & 94.43 & 61.00  & 84.50  & 34.87 & 92.67 & 81.77 & 67.70  & 91.90  & 75.30  & 70.67 & 78.47 & 75.75  \\
    {+LatHAdapter } &       & 95.00    & 63.83 & 91.67 & 39.38 & 93.63 & 87.71 & 68.63 & 92.34 & 76.45 & 74.06 & 81.60  & 78.57  \\
          $\triangle$ &        &\textcolor{blue}{(+0.57)} &\textcolor{blue}{(+2.83)} &\textcolor{blue}{(+7.17)} &\textcolor{blue}{(+4.51)} &\textcolor{blue}{(+0.96)} &\textcolor{blue}{(+5.94)} &\textcolor{blue}{(+0.93)} &\textcolor{blue}{(+0.44)} &\textcolor{blue}{(+1.15)} &\textcolor{blue}{(+3.39)} &\textcolor{blue}{(+3.13)} &\textcolor{blue}{(+2.82)} \\
    {PromptSRC} &       & 95.27 & 65.53 & 86.30  & 37.47 & 93.87 & 86.17 & 71.07 & 93.43 & 77.13 & 74.00  & 81.57 & 78.35  \\
    {+LatHAdapter } &       & 95.17 & 66.67 & 92.50  & 40.42 & 94.36 & 88.03 & 71.26 & 94.00  & 77.42 & 75.16 & 83.42 & 79.86  \\
          $\triangle$  &       &\textcolor{blue}{(-0.10)} &\textcolor{blue}{(+1.14)} &\textcolor{blue}{(+6.20)} &\textcolor{blue}{(+2.95)} &\textcolor{blue}{(+0.49)} &\textcolor{blue}{(+1.86)} &\textcolor{blue}{(+0.19)} &\textcolor{blue}{(+0.57)} &\textcolor{blue}{(+0.29)} &\textcolor{blue}{(+1.16)} &\textcolor{blue}{(+1.85)} &\textcolor{blue}{(+1.51) } \\
    \midrule
    {CoOp } & \multirow{8}[2]{*}{8} & 94.37 & 64.77 & 78.07 & 39.00  & 94.97 & 82.67 & 70.63 & 91.27 & 79.30  & 71.53 & 80.20  & 76.98   \\
    {CoCoOp} &       & 95.04 & 58.89 & 68.21 & 26.61 & 84.30  & 86.97 & 70.63 & 93.45 & 70.44 & 70.84 & 77.14 & 72.96  \\
    {GraphAdapter} & & 95.29	&67.79	&80.53	&40.02	&96.47	&87.03	&71.71	&92.15	&78.06	&74.14	&82.53 &78.70\\
    {MaPLe} &       & 95.20  & 66.50  & 87.73 & 42.00  & 95.80  & 83.60  & 70.30  & 92.57 & 79.47 & 73.23 & 81.37 & 78.89  \\
    {+LatHAdapter } &       & 95.42 & 70.22 & 92.50  & 42.83 & 96.45 & 88.36 & 71.64 & 95.94 & 79.34 & 75.31 & 84.82 & 81.17  \\
          $\triangle$ &        &\textcolor{blue}{(+0.22)} &\textcolor{blue}{(+3.72)} &\textcolor{blue}{(+4.77)} &\textcolor{blue}{(+0.83)} &\textcolor{blue}{(+0.65)} &\textcolor{blue}{(+4.76)} &\textcolor{blue}{(+1.34)} &\textcolor{blue}{(+3.37)} &\textcolor{blue}{(-0.13)} &\textcolor{blue}{(+2.08)} &\textcolor{blue}{(+3.45)} &\textcolor{blue}{(+2.28) } \\
    {PromptSRC} &       & 95.67 & 69.87 & 88.80  & 43.27 & 96.27 & 86.90  & 72.33 & 93.50  & 80.97 & 75.73  & 84.30  & 80.69  \\
    {+LatHAdapter } &       & 95.83 & 70.09 & 94.17 & 45.75 & 96.00  & 89.11 & 72.42 & 95.49 & 82.14 & 76.83 & 86.79 & 82.24  \\
          $\triangle$ &         &\textcolor{blue}{(+0.16)} &\textcolor{blue}{(+0.22)} &\textcolor{blue}{(+5.37)} &\textcolor{blue}{(+2.48)} &\textcolor{blue}{(-0.27)} &\textcolor{blue}{(+2.21)} &\textcolor{blue}{(+0.09)} &\textcolor{blue}{(+1.99)} &\textcolor{blue}{(+1.17)} &\textcolor{blue}{(+1.10)} &\textcolor{blue}{(+2.49)} &\textcolor{blue}{(+1.55) } \\
    \midrule
    {CoOp } & \multirow{8}[2]{*}{16} & 95.57 & 69.87 & 84.93 & 43.40  & 97.07 & 84.20  & 71.87 & 91.87 & 83.07 & 74.67 & 82.23 & 79.89   \\
    {CoCoOp} &       & 95.16 & 63.04 & 73.32 & 31.21 & 87.84 & 87.25 & 70.83 & 93.34 & 71.57 & 72.15 & 78.14 & 74.90  \\
    {GraphAdapter} & & 95.66	&72.40	&86.21	&44.97	&97.77	&87.27	&73.40	&92.83	&83.85	&75.19	&84.54 &81.28\\
    {MaPLe} &       & 96.00  & 71.33 & 92.33 & 48.40  & 97.00  & 85.33 & 72.33 & 92.83 & 83.57 & 75.53 & 85.03 & 81.79  \\
    {+LatHAdapter } &       & 95.75 & 74.65 & 93.33 & 47.75 & 97.55 & 88.86 & 72.71 & 97.30  & 83.29 & 77.23 & 87.95 & 83.31  \\
          $\triangle$ &         &\textcolor{blue}{(-0.25)} &\textcolor{blue}{(+3.32)} &\textcolor{blue}{(+1.00)} &\textcolor{blue}{(-0.65)} &\textcolor{blue}{(+0.55)} &\textcolor{blue}{(+3.53)} &\textcolor{blue}{(+0.38)} &\textcolor{blue}{(+4.47)} &\textcolor{blue}{(-0.28)} &\textcolor{blue}{(+1.70)} &\textcolor{blue}{(+2.92)} &\textcolor{blue}{(+1.52) } \\
    {PromptSRC} &       & 96.07 & 72.73 & 92.43 & 50.83 & 97.60  & 87.50  & 73.17 & 93.67 & 83.83 & 77.23 & 86.47 & 82.87  \\
    {+LatHAdapter} &       & 96.50  & 76.24 & 95.00    & 50.88 & 97.63 & 89.27 & 73.26 & 97.52 & 84.05 & 77.75 & 91.01 & 84.46  \\
          $\triangle$   &       &\textcolor{blue}{(+0.43)} &\textcolor{blue}{(+3.51)} &\textcolor{blue}{(+2.57)} &\textcolor{blue}{(+0.05)} &\textcolor{blue}{(+0.03)} &\textcolor{blue}{(+1.77)} &\textcolor{blue}{(+0.09)} &\textcolor{blue}{(+3.85)} &\textcolor{blue}{(+0.22)} &\textcolor{blue}{(+0.52)} &\textcolor{blue}{(+4.54)} &\textcolor{blue}{(+1.60) } \\
    \bottomrule
    \end{tabular}%
  \label{tab_few_shot}%
  }
\end{table*}%

\subsection{Ablation Experiments}
\subsubsection{Effectiveness of HHL}
To evaluate the effectiveness of Hyperbolic Hierarchical Learning (HHL), we conduct two experiments: the first uses the Attribute-aware Text Refiner (ATR) to refine textual features without introducing any additional hierarchical constraints (PromptSRC+ATR), and the second incorporates the HHL constraint into VLMs (PromptSRC+ATR+HHL).  As shown in Tab.~\ref{HiAL}, the performance of PromptSRC+ATR decreases on both base and new classes. This indicates that simply incorporating attribute prompts without explicit hierarchical constraints limits the model’s ability to learn discriminative and transferable attribute representations for the downstream training data. After integrating the HHL constraint, performance improves by 1.01\% on base classes and 1.19\% on new classes. This demonstrates the importance of modeling semantic hierarchical relationships between text classes, attribute prompts, and visual samples, enabling more robust and generalizable adapter learning.

\subsubsection{Influence of $\beta$ in ATR}
\begin{figure}[!h]
  \centering
  \includegraphics[width=1\linewidth]{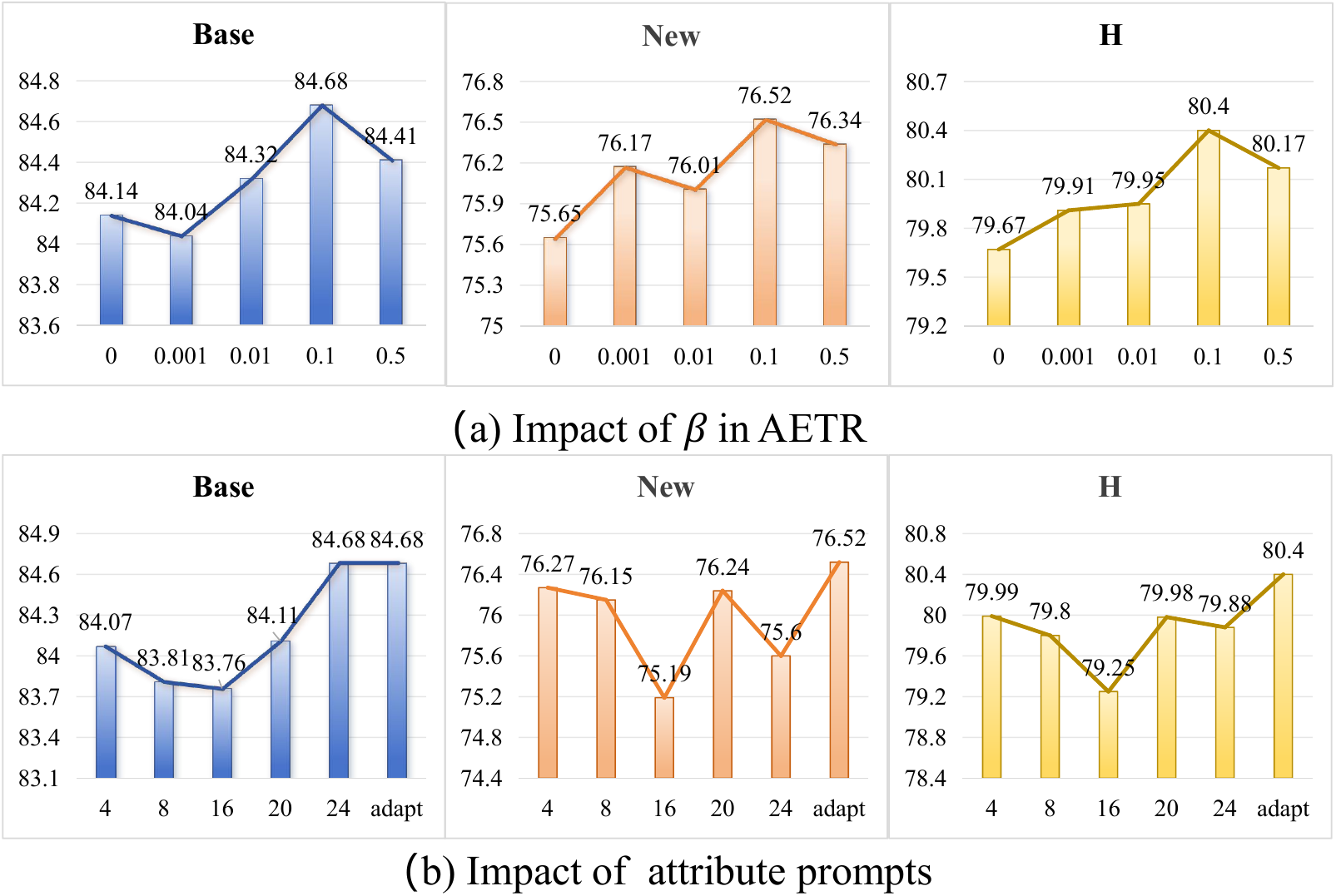}
\caption{(a) Effect of the hyperparameter $\beta$ in Attribute-aware Text Refiner (ATR) module. (b) Ablation study of the number of attribute prompts. These experiments are built on the PromptSRC+LatHAdapter framework and fine-tuned on the Base-to-New generalization task. We report the average performance across 11 datasets on both base (Base) and new classes (New), while the harmonic mean (H) provides a comprehensive metric to evaluate the performance between them.} 
\label{p_beta}
\end{figure}

To evaluate the influence of $\beta$ in the Attribute-aware Text Refiner (ATR) module, we conduct experiments based on the PromptSRC+LatHAdapter framework. In Fig.~\ref{p_beta} (left), we report the average performance across 11 datasets on the base-to-new generalization task. We achieve the highest accuracy of 84.68\% on base classes and 76.52\% on new classes when $\beta$ is set to 0.1. As $\beta$ increases from 0 to 0.1, we observe a consistent improvement in the H value. This shows that the learnable attribute prompts effectively refine textual features, thereby enhancing performance on both base and new classes. However, performance slightly decreases when $\beta$ is set to 0.5, indicating that excessive focus on attribute prompts can lead to ignoring the learning of textual features.

\subsubsection{Influence of the number of attribute prompts}
We further analyze the impact of the number of attribute prompts in Fig.~\ref{p_beta} (right). The experiments are conducted on the PromptSRC+LatHAdapter framework. Adding more prompts consistently increases base-class accuracy, but it exhibits unstable performance on new classes. This indicates that more prompts enable more fine-grained attribute learning from training samples, enhancing discrimination among base classes. However, it can also introduce irrelevant cues that influence generalization to new classes. The `adapt' denotes the optimal number of attribute prompts for each dataset. The selected prompt count is correlated with the number of classes and the challenge of the dataset. In Tab.~\ref{prompt_len}, we present the selected number of attribute prompts for each dataset. Specifically, for large-scale datasets such as ImageNet and SUN397, more prompts are beneficial, while less challenging datasets require fewer attribute prompts to achieve optimal performance.
\begin{table}[!t]
  \centering
  \caption{The number of prompts for each dataset based on the PromptSRC+LatHAdapter framework in the base-to-new tasks.}
    \begin{tabular}{c|cccc}
    \toprule
      Dataset & {ImageNet} &{Caltech101} & {OxfordPets} & {StanfordCars}  \\
    \hline
    Number & 20 &8 & 12 & 18  \\
    \midrule \midrule
    Dataset  & {Flowers102} & {Food101}& {FGVCAircraft} & {SUN397} \\
    \hline
    Number   & 14 & 16 &8 & 24  \\
    \midrule \midrule
     Dataset & {DTD} & {EuroSAT} & {UCF101}& \\
     \hline
     Number   &14 & 10 & 10 &  \\
    \bottomrule
    \end{tabular}%
  \label{prompt_len}%
\end{table}%


\subsection{Visualization Results Analysis}
\begin{figure}[!h]
  \centering
  \includegraphics[width=1\linewidth]{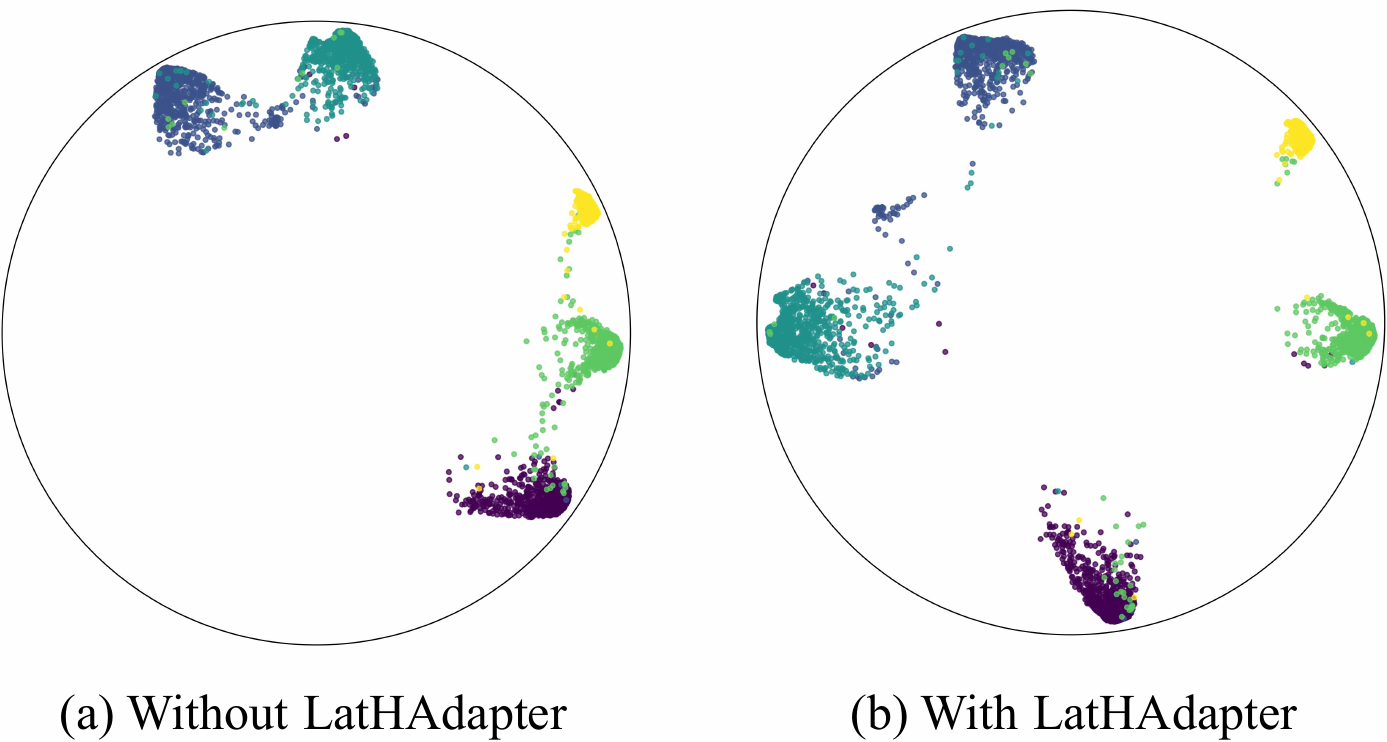}
\caption{We visualize the visual features in the hyperbolic space using the PromptSRC method on the EuroSAT dataset. (a) The distribution without integrating the LatHAdapter. (b) The distribution with integrating the LatHAdapter. } 
\label{visual}
\end{figure}

We visualize the distribution of visual features in hyperbolic space, with or without integrating the LatHAdapter. The results are demonstrated in Fig.~\ref{visual}, where each color represents a different class. Fine-grained visual features are located near the boundary. After incorporating the LatHAdapter, the feature distribution of visual samples becomes more separable. The results show that the proposed LatHAdapter effectively captures the latent semantic hierarchy for the downstream training data, providing fine-grained guidance for adapting VLMs. As a result, LatHAdapter increases inter-class distance and enhances overall performance. 

\subsection{Efficiency of LatHAdapter}
In LatHAdapter, we introduce only attribute prompts of size $n*512$ in the baseline method. The added parameters are negligible. Then, we evaluate the inference efficiency after integrating the LatHAdapter into the baseline methods, using frames per second (FPS) as the metric. The results are shown in Tab.~\ref{efficy}. All experiments are conducted on a single NVIDIA RTX 3090 GPU. Increasing the batch size to 100, the FPS remains consistent with the baseline, indicating that the proposed LatHAdapter is efficient.

\begin{table}[!t]
  \centering
  \caption{Comparison of inference efficiency after integrating LatHAdapter into the baseline methods.}
    \begin{tabular}{c|ccc|c}
    \toprule
      Method & {FPS (16)} &{FPS (32)} & {FPS (100)} & {HM} \\
    \midrule
    PromptSRC & 322 &387 & 387 & 79.67  \\
    \rowcolor{gray!20}
    PromptSRC+LatHAdapter & 193 &276 & 387 & 80.40  \\
    \midrule
    MaPLe & 315 &420 & 472 & 78.03  \\
    \rowcolor{gray!20}
    MaPLe+LatHAdapter & 291 &378 & 472 & 79.15  \\
    \bottomrule
    \end{tabular}%
  \label{efficy}%
\end{table}%

\section{Conclusion}
In this paper, we propose a novel Latent Hierarchical Adapter (LatHAdapter) for fine-tuning VLMs on downstream few-shot classification tasks. LatHAdapter first introduces learnable attribute prompts to establish an association between known and unknown classes. To enhance fine-grained textual representations and improve the generalization to new categories, we further design an Attribute-aware Text Refiner (ATR) module that exploits attribute information and task-specific classes to refine textual embeddings. Specifically, LatHAdapter employs a hierarchical regularization to learn the compact
representations and latent semantic hierarchy among categories, attributes, and images for the downstream training
data, thereby providing fine-grained guidance for adapting VLMs. The proposed LatHAdapter is a plug-and-play scheme. By integrating LatHAdapter into existing VLM fine-tuning frameworks, we achieve consistent performance improvement with negligible additional parameters.

\bibliographystyle{IEEEtran}
\bibliography{reference}

\begin{thebibliography}{10}
\providecommand{\url}[1]{#1}
\csname url@samestyle\endcsname
\providecommand{\newblock}{\relax}
\providecommand{\bibinfo}[2]{#2}
\providecommand{\BIBentrySTDinterwordspacing}{\spaceskip=0pt\relax}
\providecommand{\BIBentryALTinterwordstretchfactor}{4}
\providecommand{\BIBentryALTinterwordspacing}{\spaceskip=\fontdimen2\font plus
\BIBentryALTinterwordstretchfactor\fontdimen3\font minus \fontdimen4\font\relax}
\providecommand{\BIBforeignlanguage}[2]{{%
\expandafter\ifx\csname l@#1\endcsname\relax
\typeout{** WARNING: IEEEtran.bst: No hyphenation pattern has been}%
\typeout{** loaded for the language `#1'. Using the pattern for}%
\typeout{** the default language instead.}%
\else
\language=\csname l@#1\endcsname
\fi
#2}}
\providecommand{\BIBdecl}{\relax}
\BIBdecl

\bibitem{radford2021learning}
A.~Radford, J.~W. Kim, C.~Hallacy, P.~Ramesh, J.~Clark \emph{et~al.}, ``Learning transferable visual models from natural language supervision,'' in \emph{International conference on machine learning}, 2021, pp. 8748--8763.

\bibitem{jia2021scaling}
C.~Jia, Y.~Yang, Y.~Xia, Y.-T. Chen, Z.~Parekh, H.~Pham, Q.~Le, Y.-H. Sung, Z.~Li, and T.~Duerig, ``Scaling up visual and vision-language representation learning with noisy text supervision,'' in \emph{International conference on machine learning}, 2021, pp. 4904--4916.

\bibitem{hu2024comma}
L.~Hu, L.~Gao, Z.~Liu, C.-M. Pun, and W.~Feng, ``Comma: Co-articulated multi-modal learning,'' in \emph{Proceedings of the AAAI Conference on Artificial Intelligence}, vol.~38, no.~3, 2024, pp. 2238--2246.

\bibitem{khattak2023self}
M.~U. Khattak, S.~T. Wasim, M.~Naseer, S.~Khan, M.-H. Yang, and F.~S. Khan, ``Self-regulating prompts: Foundational model adaptation without forgetting,'' in \emph{Proceedings of the IEEE International Conference on Computer Vision}, 2023, pp. 15\,190--15\,200.

\bibitem{jia2022visual}
M.~Jia, L.~Tang, B.-C. Chen, C.~Cardie, S.~Belongie, B.~Hariharan, and S.-N. Lim, ``Visual prompt tuning,'' in \emph{European Conference on Computer Vision}, 2022, pp. 709--727.

\bibitem{zhou2022learning}
K.~Zhou, J.~Yang, C.~C. Loy, and Z.~Liu, ``Learning to prompt for vision-language models,'' \emph{International Journal of Computer Vision}, vol. 130, no.~9, pp. 2337--2348, 2022.

\bibitem{hayou2024lora}
S.~Hayou, N.~Ghosh, and B.~Yu, ``Lora+: Efficient low rank adaptation of large models,'' \emph{arXiv preprint arXiv:2402.12354}, 2024.

\bibitem{hu2022lora}
E.~J. Hu, Y.~Shen, P.~Wallis, Z.~Allen-Zhu, Y.~Li, S.~Wang, L.~Wang, W.~Chen \emph{et~al.}, ``Lora: Low-rank adaptation of large language models.'' \emph{ICLR}, vol.~1, no.~2, p.~3, 2022.

\bibitem{valipour2022dylora}
M.~Valipour, M.~Rezagholizadeh, I.~Kobyzev, and A.~Ghodsi, ``Dylora: Parameter efficient tuning of pre-trained models using dynamic search-free low-rank adaptation,'' \emph{arXiv preprint arXiv:2210.07558}, 2022.

\bibitem{shi2024reslora}
S.~Shi, S.~Huang, M.~Song, Z.~Li, Z.~Zhang, H.~Huang, F.~Wei, W.~Deng, F.~Sun, and Q.~Zhang, ``Reslora: Identity residual mapping in low-rank adaption,'' \emph{arXiv preprint arXiv:2402.18039}, 2024.

\bibitem{zhang2022tip}
R.~Zhang, W.~Zhang, R.~Fang, P.~Gao, K.~Li, J.~Dai, Y.~Qiao, and H.~Li, ``Tip-adapter: Training-free adaption of clip for few-shot classification,'' in \emph{European conference on computer vision}, 2022, pp. 493--510.

\bibitem{yu2023task}
T.~Yu, Z.~Lu, X.~Jin, Z.~Chen, and X.~Wang, ``Task residual for tuning vision-language models,'' in \emph{Proceedings of the IEEE Conference on Computer Vision and Pattern Recognition}, 2023, pp. 10\,899--10\,909.

\bibitem{gao2024clip}
P.~Gao, S.~Geng, R.~Zhang, T.~Ma, R.~Fang, Y.~Zhang, H.~Li, and Y.~Qiao, ``Clip-adapter: Better vision-language models with feature adapters,'' \emph{International Journal of Computer Vision}, vol. 132, no.~2, pp. 581--595, 2024.

\bibitem{li2024graphadapter}
X.~Li, D.~Lian, Z.~Lu, J.~Bai, Z.~Chen, and X.~Wang, ``Graphadapter: Tuning vision-language models with dual knowledge graph,'' \emph{Advances in Neural Information Processing Systems}, vol.~36, 2024.

\bibitem{khattak2023maple}
M.~U. Khattak, H.~Rasheed, M.~Maaz, S.~Khan, and F.~S. Khan, ``Maple: Multi-modal prompt learning,'' in \emph{Proceedings of the IEEE Conference on Computer Vision and Pattern Recognition}, 2023, pp. 19\,113--19\,122.

\bibitem{si2024unleashing}
C.~Si, Z.~Shi, S.~Zhang, X.~Yang, H.~Pfister, and W.~Shen, ``Unleashing the power of task-specific directions in parameter efficient fine-tuning,'' in \emph{The Thirteenth International Conference on Learning Representations}, 2024.

\bibitem{seputis2024multi}
D.~Seputis, S.~Mihailov, S.~Chatterjee, and Z.~Xiao, ``Multi-modal adapter for vision-language models,'' in \emph{Proceedings of the IEEE Conference on Computer Vision and Pattern Recognition}, 2024.

\bibitem{hao2025task}
F.~Hao, F.~He, F.~Wu, T.~Wang, C.~Song, and J.~Cheng, ``Task-aware clustering for prompting vision-language models,'' in \emph{Proceedings of the Computer Vision and Pattern Recognition Conference}, 2025, pp. 14\,745--14\,755.

\bibitem{li2025atprompt}
Z.~Li, Y.~Song, P.~Zhao, M.-M. Cheng, X.~Li, and J.~Yang, ``Textual prompt learning with embedded attributes,'' in \emph{Proceedings of the IEEE International Conference on Computer Vision}, 2025.

\bibitem{desai2023hyperbolic}
K.~Desai, M.~Nickel, T.~Rajpurohit, J.~Johnson, and S.~R. Vedantam, ``Hyperbolic image-text representations,'' in \emph{International Conference on Machine Learning}.\hskip 1em plus 0.5em minus 0.4em\relax PMLR, 2023, pp. 7694--7731.

\bibitem{ramasinghe2024accept}
S.~Ramasinghe, V.~Shevchenko, G.~Avraham, and A.~Thalaiyasingam, ``Accept the modality gap: An exploration in the hyperbolic space,'' in \emph{Proceedings of the IEEE Conference on Computer Vision and Pattern Recognition}, 2024, pp. 27\,263--27\,272.

\bibitem{tang2024amu}
Y.~Tang, Z.~Lin, Q.~Wang, P.~Zhu, and Q.~Hu, ``Amu-tuning: Effective logit bias for clip-based few-shot learning,'' in \emph{Proceedings of the IEEE Conference on Computer Vision and Pattern Recognition}, 2024, pp. 23\,323--23\,333.

\bibitem{kim2023hier}
S.~Kim, B.~Jeong, and S.~Kwak, ``Hier: Metric learning beyond class labels via hierarchical regularization,'' in \emph{Proceedings of the IEEE conference on computer vision and pattern recognition}, 2023, pp. 19\,903--19\,912.

\bibitem{Compositional2024}
C.~E.~L. for Hyperbolic Vision-Language~Models, ``Adaptive budget allocation for parameter-efficient finetuning,'' \emph{arXiv preprint arXiv:2410.06912}, 2024.

\bibitem{Open2025}
Z.~Peng, Z.~Xu, Z.~Zeng, C.~Wen, Y.~Huang, M.~Yang, F.~Tang, and W.~Shen, ``Understanding fine-tuning clip for open-vocabulary semantic segmentation in hyperbolic space,'' in \emph{Proceedings of the Computer Vision and Pattern Recognition Conference}, 2025, pp. 4562--4572.

\bibitem{poppi2025hyperbolic}
T.~Poppi, T.~Kasarla, P.~Mettes, L.~Baraldi, and R.~Cucchiara, ``Hyperbolic safety-aware vision-language models,'' in \emph{Proceedings of the Computer Vision and Pattern Recognition Conference}, 2025, pp. 4222--4232.

\bibitem{Chen_2021}
C.~Jia, Y.~Yang, Y.~Xia, Y.-T. Chen, Z.~Parekh, H.~Pham, Q.~Le, Y.-H. Sung, Z.~Li, and T.~Duerig, ``Scaling up visual and vision-language representation learning with noisy text supervision,'' in \emph{International conference on machine learning}, 2021, pp. 4904--4916.

\bibitem{Vilbert2019}
J.~Lu, D.~Batra, D.~Parikh, and S.~Lee, ``Vilbert: Pretraining task-agnostic visiolinguistic representations for vision-and-language tasks,'' \emph{Advances in neural information processing systems}, vol.~32, no.~2, 2019.

\bibitem{Conceptual2024}
Y.~Zhang, K.~Yu, S.~Wu, and Z.~He, ``Conceptual codebook learning for vision-language models,'' \emph{arXiv preprint arXiv:2407.02350}, 2024.

\bibitem{zhou2022conditional}
K.~Zhou, J.~Yang, C.~C. Loy, and Z.~Liu, ``Conditional prompt learning for vision-language models,'' in \emph{Proceedings of the IEEE conference on computer vision and pattern recognition}, 2022, pp. 16\,816--16\,825.

\bibitem{xie2024textrefiner}
J.~Xie, Y.~Zhang, J.~Peng, Z.~Huang, and L.~Cao, ``Textrefiner: Internal visual feature as efficient refiner for vision-language models prompt tuning,'' pp. 8718--8726, 2025.

\bibitem{li2025dpc}
H.~Li, L.~Wang, C.~Wang, J.~Jiang, Y.~Peng, and G.~Long, ``Dpc: Dual-prompt collaboration for tuning vision-language models,'' in \emph{Proceedings of the Computer Vision and Pattern Recognition Conference}, 2025, pp. 25\,623--25\,632.

\bibitem{zhao2024hegraphadapter}
Y.~Zhao, B.~Jiang, X.~Wang, Q.~Xu, and J.~Tang, ``Hegraphadapter: Tuning multi-modal vision-language models with heterogeneous graph adapter,'' \emph{arXiv preprint arXiv:2410.07854}, 2024.

\bibitem{mao2024dora}
Y.~Mao, K.~Huang, C.~Guan, G.~Bao, F.~Mo, and J.~Xu, ``Dora: Enhancing parameter-efficient fine-tuning with dynamic rank distribution,'' \emph{arXiv preprint arXiv:2405.17357}, 2024.

\bibitem{zhang2023adalora}
Q.~Zhang, M.~Chen, A.~Bukharin, N.~Karampatziakis, P.~He, Y.~Cheng, W.~Chen, and T.~Zhao, ``Adalora: Adaptive budget allocation for parameter-efficient fine-tuning,'' \emph{arXiv preprint arXiv:2303.10512}, 2023.

\bibitem{Proto2024}
J.~J. P. K. P. Y.-W. C. X. D.~Y. Xiang, ``Proto-clip: Vision-language prototypical network for few-shot learning,'' \emph{IEEE International Conference on Intelligent Robots and Systems}, 2024.

\bibitem{guo2025mmrl}
Y.~Guo and X.~Gu, ``Mmrl: Multi-modal representation learning for vision-language models,'' in \emph{Proceedings of the Computer Vision and Pattern Recognition Conference}, 2025, pp. 25\,015--25\,025.

\bibitem{song2023meta}
L.~Song, R.~Xue, H.~Wang, H.~Sun, Y.~Ge, Y.~Shan \emph{et~al.}, ``Meta-adapter: An online few-shot learner for vision-language model,'' \emph{Advances in Neural Information Processing Systems}, vol.~36, pp. 55\,361--55\,374, 2023.

\bibitem{2025Hyperbolic}
X.~Yang, D.~Kong, N.~Wang, and X.~Gao, ``Hyperbolic insights with knowledge distillation for cross-domain few-shot learning,'' \emph{IEEE Transactions on Image Processing}, 2025.

\bibitem{2021Curvature}
Z.~Gao, Y.~Wu, Y.~Jia, and M.~Harandi, ``Curvature generation in curved spaces for few-shot learning,'' in \emph{Proceedings of the IEEE international conference on computer vision}, 2021, pp. 8691--8700.

\bibitem{krioukov2010hyperbolic}
D.~Krioukov, F.~Papadopoulos, M.~Kitsak, A.~Vahdat, and M.~Bogun{\'a}, ``Hyperbolic geometry of complex networks,'' \emph{Physical Review E—Statistical, Nonlinear, and Soft Matter Physics}, vol.~82, no.~3, p. 036106, 2010.

\bibitem{cannon1997hyperbolic}
J.~W. Cannon, W.~J. Floyd, R.~Kenyon, W.~R. Parry \emph{et~al.}, ``Hyperbolic geometry,'' \emph{Flavors of geometry}, vol.~31, no. 59-115, p.~2, 1997.

\bibitem{nickel2017poincare}
M.~Nickel and D.~Kiela, ``Poincar{\'e} embeddings for learning hierarchical representations,'' \emph{Advances in neural information processing systems}, vol.~30, 2017.

\bibitem{ganea2018hyperbolic}
O.~Ganea, G.~B{\'e}cigneul, and T.~Hofmann, ``Hyperbolic neural networks,'' \emph{Advances in neural information processing systems}, vol.~31, 2018.

\bibitem{nickel2018learning}
M.~Nickel and D.~Kiela, ``Learning continuous hierarchies in the lorentz model of hyperbolic geometry,'' in \emph{International conference on machine learning}, 2018, pp. 3779--3788.

\bibitem{hu2024rethinking}
C.~Hu, K.-Y. Zhang, T.~Yao, S.~Ding, and L.~Ma, ``Rethinking generalizable face anti-spoofing via hierarchical prototype-guided distribution refinement in hyperbolic space,'' in \emph{Proceedings of the IEEE Conference on Computer Vision and Pattern Recognition}, 2024, pp. 1032--1041.

\bibitem{sarkar2011low}
R.~Sarkar, ``Low distortion delaunay embedding of trees in hyperbolic plane,'' in \emph{International symposium on graph drawing}.\hskip 1em plus 0.5em minus 0.4em\relax Springer, 2011, pp. 355--366.

\bibitem{Abraham2008}
A.~A. Ungar, ``A gyrovector space approach to hyperbolic geometry,'' vol.~3, pp. 1--194, 2008.

\bibitem{liang2022mind}
W.~Liang, Y.~Zhang, Y.~Kwon, S.~Yeung, and J.~Zou, ``Mind the gap: Understanding the modality gap in multi-modal contrastive representation learning,'' \emph{arXiv preprint arXiv:2203.02053}, 2022.

\bibitem{yang2024hyperbolic}
M.~Yang, A.~Feng, B.~Xiong, J.~Liu, I.~King, and R.~Ying, ``Hyperbolic fine-tuning for large language models,'' \emph{arXiv preprint arXiv:2410.04010}, 2024.

\bibitem{kong2024hyperbolic}
F.~Kong, Y.~Chen, J.~Cai, and D.~Modolo, ``Hyperbolic learning with synthetic captions for open-world detection,'' in \emph{Proceedings of the IEEE Conference on Computer Vision and Pattern Recognition}, 2024, pp. 16\,762--16\,771.

\bibitem{recht2019imagenet}
B.~Recht, R.~Roelofs, L.~Schmidt, and V.~Shankar, ``Do imagenet classifiers generalize to imagenet?'' in \emph{International conference on machine learning}, 2019, pp. 5389--5400.

\bibitem{wang2019learning}
H.~Wang, S.~Ge, Z.~Lipton, and E.~P. Xing, ``Learning robust global representations by penalizing local predictive power,'' \emph{Advances in Neural Information Processing Systems}, vol.~32, 2019.

\bibitem{hendrycks2021natural}
D.~Hendrycks, K.~Zhao, S.~Basart, J.~Steinhardt, and D.~Song, ``Natural adversarial examples,'' in \emph{Proceedings of the IEEE Conference on Computer Vision and Pattern Recognition}, 2021, pp. 15\,262--15\,271.

\bibitem{hendrycks2021many}
D.~Hendrycks, S.~Basart, N.~Mu, S.~Kadavath, F.~Wang, E.~Dorundo, R.~Desai, T.~Zhu, S.~Parajuli, M.~Guo \emph{et~al.}, ``The many faces of robustness: A critical analysis of out-of-distribution generalization,'' in \emph{Proceedings of the IEEE Conference on Computer Vision and Pattern Recognition}, 2021, pp. 8340--8349.

\bibitem{deng2009imagenet}
J.~Deng, W.~Dong, R.~Socher, L.-J. Li, K.~Li, and L.~Fei-Fei, ``Imagenet: A large-scale hierarchical image database,'' in \emph{2009 IEEE conference on computer vision and pattern recognition}, 2009, pp. 248--255.

\bibitem{fei2004learning}
L.~Fei-Fei, R.~Fergus, and P.~Perona, ``Learning generative visual models from few training examples: An incremental bayesian approach tested on 101 object categories,'' in \emph{2004 conference on computer vision and pattern recognition workshop}, 2004, pp. 178--178.

\bibitem{parkhi2012cats}
O.~M. Parkhi, A.~Vedaldi, A.~Zisserman, and C.~Jawahar, ``Cats and dogs,'' in \emph{2012 IEEE conference on computer vision and pattern recognition}, 2012, pp. 3498--3505.

\bibitem{krause20133d}
J.~Krause, M.~Stark, J.~Deng, and L.~Fei-Fei, ``3d object representations for fine-grained categorization,'' in \emph{Proceedings of the IEEE Conference on Computer Vision and Pattern Recognition workshops}, 2013, pp. 554--561.

\bibitem{nilsback2008automated}
M.-E. Nilsback and A.~Zisserman, ``Automated flower classification over a large number of classes,'' in \emph{2008 Sixth Indian conference on computer vision, graphics \& image processing}, 2008, pp. 722--729.

\bibitem{bossard2014food}
L.~Bossard, M.~Guillaumin, and L.~Van~Gool, ``Food-101--mining discriminative components with random forests,'' in \emph{Proceedings of the IEEE Conference on Computer vision ECCV 2014}, 2014, pp. 446--461.

\bibitem{maji2013fine}
S.~Maji, E.~Rahtu, J.~Kannala, M.~Blaschko, and A.~Vedaldi, ``Fine-grained visual classification of aircraft,'' \emph{arXiv preprint arXiv:1306.5151}, 2013.

\bibitem{xiao2010sun}
J.~Xiao, J.~Hays, K.~A. Ehinger, A.~Oliva, and A.~Torralba, ``Sun database: Large-scale scene recognition from abbey to zoo,'' in \emph{2010 IEEE computer society conference on computer vision and pattern recognition}, 2010, pp. 3485--3492.

\bibitem{cimpoi2014describing}
M.~Cimpoi, S.~Maji, I.~Kokkinos, S.~Mohamed, and A.~Vedaldi, ``Describing textures in the wild,'' in \emph{Proceedings of the IEEE Conference on Computer Vision and Pattern Recognition}, 2014, pp. 3606--3613.

\bibitem{helber2019eurosat}
P.~Helber, B.~Bischke, A.~Dengel, and D.~Borth, ``Eurosat: A novel dataset and deep learning benchmark for land use and land cover classification,'' \emph{IEEE Journal of Selected Topics in Applied Earth Observations and Remote Sensing}, vol.~12, no.~7, pp. 2217--2226, 2019.

\bibitem{soomro2012ucf101}
K.~Soomro, A.~R. Zamir, and M.~Shah, ``Ucf101: A dataset of 101 human actions classes from videos in the wild,'' \emph{arXiv preprint arXiv:1212.0402}, 2012.

\bibitem{li2024promptkd}
Z.~Li, X.~Li, X.~Fu, X.~Zhang, W.~Wang, S.~Chen, and J.~Yang, ``Promptkd: Unsupervised prompt distillation for vision-language models,'' in \emph{Proceedings of the IEEE Conference on Computer Vision and Pattern Recognition}, 2024, pp. 26\,617--26\,626.

\bibitem{zhang2024dept}
J.~Zhang, S.~Wu, L.~Gao, H.~T. Shen, and J.~Song, ``Dept: Decoupled prompt tuning,'' in \emph{Proceedings of the IEEE Conference on Computer Vision and Pattern Recognition}, 2024, pp. 12\,924--12\,933.

\bibitem{yao2024tcp}
H.~Yao, R.~Zhang, and C.~Xu, ``Tcp: Textual-based class-aware prompt tuning for visual-language model,'' in \emph{Proceedings of the IEEE Conference on Computer Vision and Pattern Recognition}, 2024, pp. 23\,438--23\,448.

\end{thebibliography}

\begin{IEEEbiographynophoto}{Yumiao Zhao}
is currently a Ph.D. student in Computer Science and Technology at Anhui University, Hefei, China. Her research interests include face recognition, few-shot learning, and multi-modal learning.
\end{IEEEbiographynophoto}

\begin{IEEEbiographynophoto}{Bo Jiang}
received the B.S. degree in mathematics and applied mathematics, and the M.Eng. and Ph.D. degrees in computer science from Anhui University, Hefei, China, in 2009, 2012, and 2015, respectively. He is currently a Professor of computer science with Anhui University. His research interests include image matching, graph data representation and learning, and few-shot learning. He has authored or coauthored more than 100 papers, including 40 papers in top conferences and journals, such as CVPR, AAAI, MICCAI, IJCV, IEEE TPAMI, IEEE TIP, and IEEE TNNLS. He gained some research projects of NSFC and Natural Science Foundation of Anhui Province. He was the recipient of the rising star award of ACM Hefei.
\end{IEEEbiographynophoto}

\begin{IEEEbiographynophoto}{Yuhe Ding} is currently a Ph.D. student in the School of Computer Science and Technology, Anhui University, Hefei, China. 
She received her B.Eng. degree from Anhui University in July 2019.
From 2020 to 2022, she was a visiting student at CASIA.
Her research interests focus on transfer learning, pattern recognition, and computer vision.
\end{IEEEbiographynophoto}

\begin{IEEEbiographynophoto}{Xiao Wang}
received the B.S. degree from West Anhui University, Luan, China, in 2013, and the Ph.D. degree in computer science from Anhui University, Hefei, China, in 2019. From 2015 and 2016, he was a visiting student with the School of Data and Computer Science, Sun Yat-sen University, Guangzhou, China, and supervised by Professor Liang Lin. He visited UBTECH Sydney Artificial Intelligence Centre, the Faculty of Engineering, The University of Sydney, Sydney, NSW, Australia, in 2019, and supervised by Professor Dacheng Tao. He finished the Postdoctoral Research (from April 2020 to April 2022) with Pengcheng Laboratory, Shenzhen, China. Currently, he is working at the School of Computer Science and Technology of Anhui University, Hefei 230601, China, as an Associate Professor. His current research interests are mainly computer vision, event-based vision, machine learning, and pattern recognition. Dr. Wang serves as an Associate Editor for the journal "IEEE Sensors Journal", and also a reviewer for a number of journals and conferences, such as the International Journal of Computer Vision, IEEE TRANSACTIONS ON CIRCUITS AND SYSTEMS FOR VIDEO TECHNOLOGY, IEEE TRANSACTIONS ON NEURAL NETWORKS AND LEARNING SYSTEMS, IEEE TRANSACTIONS ON IMAGE PROCESSING, Computer Vision and Image Understanding, CVPR, ICCV, ECCV, AAAI, ACCV, ACM-MM, and WACV.
\end{IEEEbiographynophoto}

\begin{IEEEbiographynophoto}{Jin Tang}
received the B.Eng. degree from the School of Automation in 1999 and the Ph.D. degree from the School of Computer Science and Technology, Anhui University, Hefei, China, in 2007. He is currently a Professor with the School of Computer Science and Technology, Anhui University. He has published 30 papers in top-tier conferences and leading journals, including IEEE TIP, IEEE TCSVT, PR, CVPR, NIPS, ACM MM, and IJCAI. He has obtained 20 competitive external grants from NSFC, Anhui Province, and other commercial sources as a PI, a Co-PI, or a Senior Personnel.
His current research focuses on image processing, pattern recognition and computer vision.
\end{IEEEbiographynophoto}

\begin{IEEEbiographynophoto}{Bin Luo}
received the B.Eng. degree in electronics and M.Eng. degree in computer science from Anhui University, Hefei, China, in 1984 and 1991, respectively, and the Ph.D. degree in computer science from the University of York, York, U.K., in 2002. He is currently a Professor with Anhui University. He has authored or coauthored more than 200 papers in journal and refereed conferences. He is the Chair of the IEEE Hefei Subsection. He was a peer reviewer of international academic journals, such as IEEE Transactions on Pattern Analysis and Machine Intelligence, Pattern Recognition, and Pattern Recognition Letters. His research interests include random graph-based pattern recognition, image and graph matching, and spectral analysis.
\end{IEEEbiographynophoto}

\end{document}